\documentclass[10pt,twocolumn,letterpaper]{article}

\usepackage{cvpr}
\usepackage{times}
\usepackage{epsfig}
\usepackage{graphicx}
\usepackage{comment}
\usepackage{amsmath,amssymb} 
\usepackage{color}
\usepackage{subcaption}
\usepackage{mathrsfs}
\usepackage{array,multirow,adjustbox}
\usepackage{tabularx}
\usepackage{cite}

\usepackage[pagebackref=true,breaklinks=true,colorlinks,bookmarks=false]{hyperref}
\cvprfinalcopy 

\def\cvprPaperID{4882} 

\ifcvprfinal\fi
\begin{document}

\title{Image Restoration for Under-Display Camera}

\author{
Yuqian Zhou$^1$, David Ren$^{2}$, Neil Emerton$^3$, Sehoon Lim$^3$, Timothy Large$^3$ \\
$^1$IFP, UIUC,~~ $^2$CIL, UC Berkeley,~~ $^3$Microsoft
}

\maketitle

\begin{abstract}
The new trend of full-screen devices encourages us to position a camera behind a screen. Removing the bezel and centralizing the camera under the screen brings larger display-to-body ratio and enhances eye contact in video chat, but also causes image degradation. In this paper, we focus on a newly-defined Under-Display Camera (UDC), as a novel real-world single image restoration problem. First, we take a 4k Transparent OLED (T-OLED) and a phone Pentile OLED (P-OLED) and analyze their optical systems to understand the degradation. Second, we design a Monitor-Camera Imaging System (MCIS) for easier real pair data acquisition, and a model-based data synthesizing pipeline to generate Point Spread Function (PSF) and UDC data only from display pattern and camera measurements. Finally, we resolve the complicated degradation using deconvolution-based pipeline and learning-based methods. Our model demonstrates a real-time high-quality restoration. The presented methods and results reveal the promising research values and directions of UDC. 

\end{abstract}

\section{Introduction}
\begin{figure}[t]\setlength{\belowcaptionskip}{-10pt}
	\centering
		\includegraphics[width=0.8\linewidth]{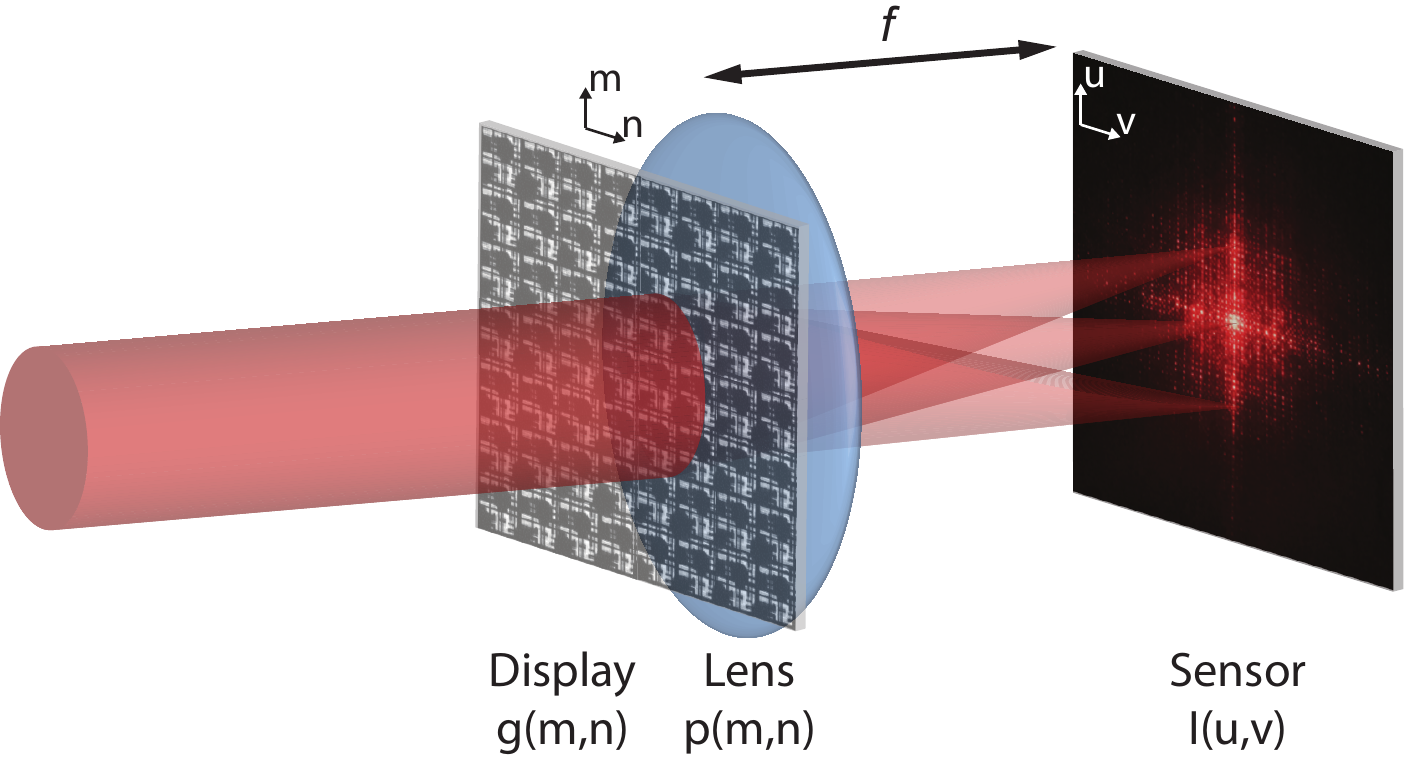}
	\caption{The newly proposed imaging system named Under-Display Camera (UDC). We mount display screen on top of a traditional digital camera lens. The design brings new trend of full-screen devices.}
	\label{fig:schematic_3d}
\end{figure}

Under-display Camera (UDC) is a new imaging system that mounts display screen on top of a traditional digital camera lens, as shown in Figure ~\ref{fig:schematic_3d}. Such a system has mainly two advantages. First, it brings a new product trend of full-screen devices~\cite{chen2019full} with larger screen-to-body ratio, which can provide better user perceptive and intelligent experience~\cite{evans2019optical}. Without seeing the bezel and extra buttons, users can easily access more functions by directly touching the screen. Second, it provides better human computer interaction. By putting the camera in the center of the display, it enhances teleconferencing experiences with perfect gaze tracking, and it is increasingly relevant for larger display devices such as laptops and TVs. 

Unlike pressure or fingerprint sensors that can be easily integrated into a display, it is relatively difficult to retain full functionality of an imaging sensor after mounting it behind a display. The imaging quality of a camera will be severely degraded due to lower light transmission rate and diffraction effects. As a result, images captured will be noisy and blurry. Therefore, while bringing better user experience and interaction, UDC may sacrifice the quality of photography, face processing \cite{tan2018face} and other downstream vision tasks. Restoring and enhancing the images captured by UDC system will be desired. 

Traditional image restoration approaches form the task as an inverse problem or an optimization problem like Maximum-a-Posterior (MAP). For the UDC problem, for practical purposes, the proposed image restoration algorithm and system are expected to work in real-time. Therefore, deconvolutional-based methods like Wiener Filter \cite{goldstein1998multistage} should be preferred. Deconvolution is the inverse process of convolution and recovers the original signal from the point-spread-function (PSF)-convolved image. The fidelity of the deconvolution process is dependent on the space-invariance of the PSF over the image field of-view (FOV) and on a low condition number for the inverse of
the PSF \cite{heath2018scientific}. For strongly non-delta-function-like PSFs such as those encountered when imaging through a display, the value of
condition number can be large. For such PSFs an additional denoising step may be essential.

Another option is the emerging discriminative learning-based image restoration model. Data-driven discriminative learning-based image restoration models usually outperform traditional methods in specific tasks like image de-noising \cite{zhang2017beyond,zhou2019awgn,zhou2019adaptation,liu2019learning,zhang2018ffdnet,abdelhamed2019ntire}, de-bluring \cite{kupyn2018deblurgan,nah2019ntire},de-raining \cite{zhang2019image,zhang2018density}, de-hazing \cite{fattal2008single,ren2016single}, super-resolution \cite{lim2017enhanced,wang2018esrgan}, and light-enhancement \cite{chen2018learning}. However, working on synthesis data with single degradation type, existing models can be hardly utilized to enhance real-world low-quality images with complicated or combined degradation types. To address complicated real degradation like the UDC problem, directly collecting real paired data or synthesizing near-realistic data after fully understanding the degradation model is necessary. 

In this paper, we present the first study to define and analyze the novel Under-Display Camera (UDC) system from both optics and image restoration viewpoints. For optics, we parse the optical system of the UDC pipeline and analyze the characteristics of light transmission. Then we relate the obtained intuitions and measurements to an image restoration pipeline, and propose two ways of resolving the single-image restoration: A deconvolution-based Wiener Filter \cite{orieux2010bayesian} pipeline (DeP) and a data-driven learning-based approach. Specifically, we regard UDC restoration as a combination of tasks such as low-light enhancement, de-blurring, and de-noising. 

Without loss of generality, our analysis focuses on \textbf{two types of displays}, a 4K Transparent Organic Light-Emitting Diode (T-OLED) and a phone Pentile OLED (P-OLED), and \textbf{a single camera type,} a 2K FLIR RGB Point Grey research camera. To obtain the real imaging data and measure the optical factors of the system, we also propose a data acquisition system using the above optical elements. 

In summary, the main contributions of our paper are: (1) A brand new imaging system named Under-Display Camera (UDC) is defined, measured and analyzed. Extensive experiments reveal the image degradation process of the system, inspiring better approaches for restoring the captured images. (2) As a baseline, two practical and potential solutions are proposed, including conventional Wiener Filter and the recent learning-based method. (3) Adopting the newly-assembled image acquisition system, we collect the first Under-Display Camera (UDC) dataset which will be released and evaluated by the public.


\section{Related Work}
\paragraph{Real-world Image Reconstruction and Restoration}
Image restoration for UDC \cite{zhang2020image,lim202074,lim2020aperture,zhou2020udc} can be categorized into the problem of Real-world restoration\cite{abdelhamed2019ntire,zhang2019zoom}. It is becoming a new concept in low-level vision. In the past decades, low-level vision works on synthetic data (denoising on AWGN and SR on Bicubic), but the models are not efficient for images with real degradation such as real noises or blur kernels. Making models perform better on real-world inputs usually requires new problem analysis and a more challenging data collection. Recently, researchers also worked on challenging cases like lensless imaging problems \cite{peng2019learned, monakhova2019learned, khan2019towards}, or integrating optics theory with High Dynamic Range imaging \cite{sun2020learning}. Previously, there has been two common ways to prepare adaptive training data for real-world problems: real data collection and near-realistic data synthesis.

Recently, more real noise datasets such as DND \cite{plotz2017benchmarking}, SIDD \cite{abdelhamed2018high,nah2019ntire}, and RENOIR \cite{anaya2018renoir}, have been introduced to address practical denoising problems. Abdelrahman et al. \cite{abdelhamed2019ntire} proposed to estimate ground truth from captured smartphone noise images, and utilized the paired data to train and evaluate the real denoising algorithms. In addition to noise, Chen et al. first introduced the SID dataset \cite{chen2018learning} to resolve extreme low-light imaging. In the area of Single Image Super Resolution (SISR), researchers considered collecting optical zoom data \cite{zhang2019zoom,chen2019camera} to learn better computational zoom. Other restoration problems including reflection removal \cite{wan2017benchmarking,punnappurath2019reflection} also follow the trend of real data acquisition. Collecting real data suffers from limitation of scene variety since most previous models acquire images of postcards, static objects or color boards. In this paper, we propose a novel monitor-camera imaging system, to add real degradation to the existing natural image datasets like DIV2K \cite{agustsson2017ntire}. 

A realistic dataset can be synthesized if the degradation model is fully understood and resolved. One good practice of data synthesis is generating real noises on raw sensors or RGB images. CBDNet \cite{guo2019toward} and Tim et al. \cite{brooks2019unprocessing} synthesized realistic noise by unfolding the in-camera pipeline, and Abdelhamed et al. \cite{abdelhamed2019noise} better fitted the real noise distribution with flow-based generative models. Zhou et al. \cite{zhou2019awgn} adapted the AWGN-RVIN noise into real RGB noise by analyzing the demosacing process. Other physics-based synthesis was also explored in blur\cite{brooks2019learning} or hazing\cite{ancuti2019dense}. For the UDC problem in this paper, we either collected real paired data, or synthesized near-realistic data from model simulation. In particular, we applied the theory of Fourier optics to simulate the diffraction effects, and further adjusted the data with other camera measurements. Our data synthesizing pipeline demonstrates a promising performance for addressing real complicated degradation problem.

\begin{figure*}[t]\setlength{\abovecaptionskip}{0pt}
	\begin{subfigure}{0.4\linewidth} 
		\includegraphics[width=\textwidth, height=5cm]{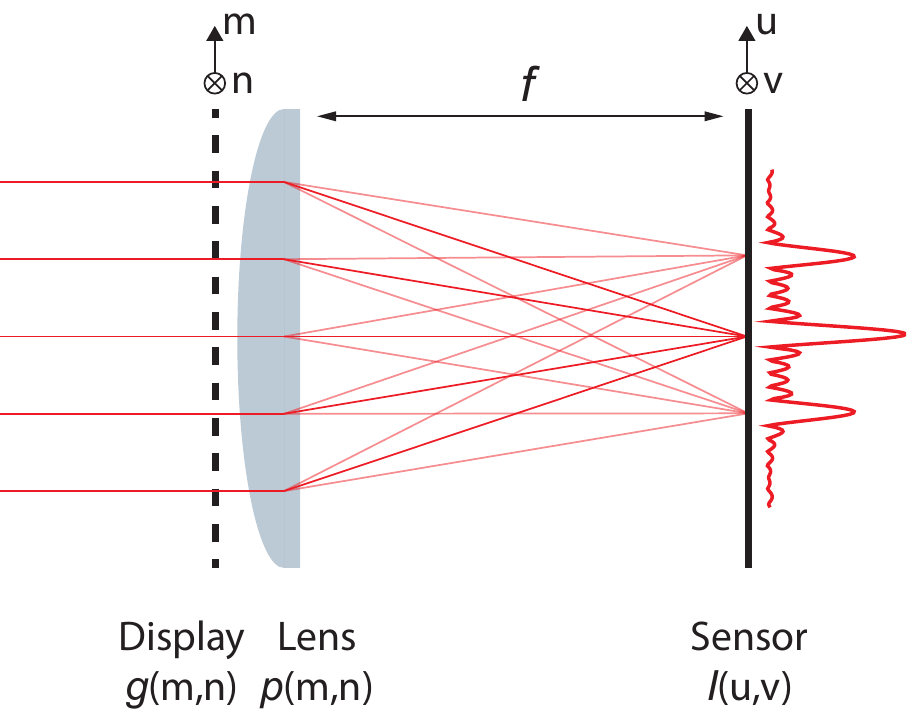}
		\caption{} 
	\end{subfigure}
	\begin{subfigure}{0.6\linewidth} 
		\includegraphics[width=\textwidth, height=5cm]{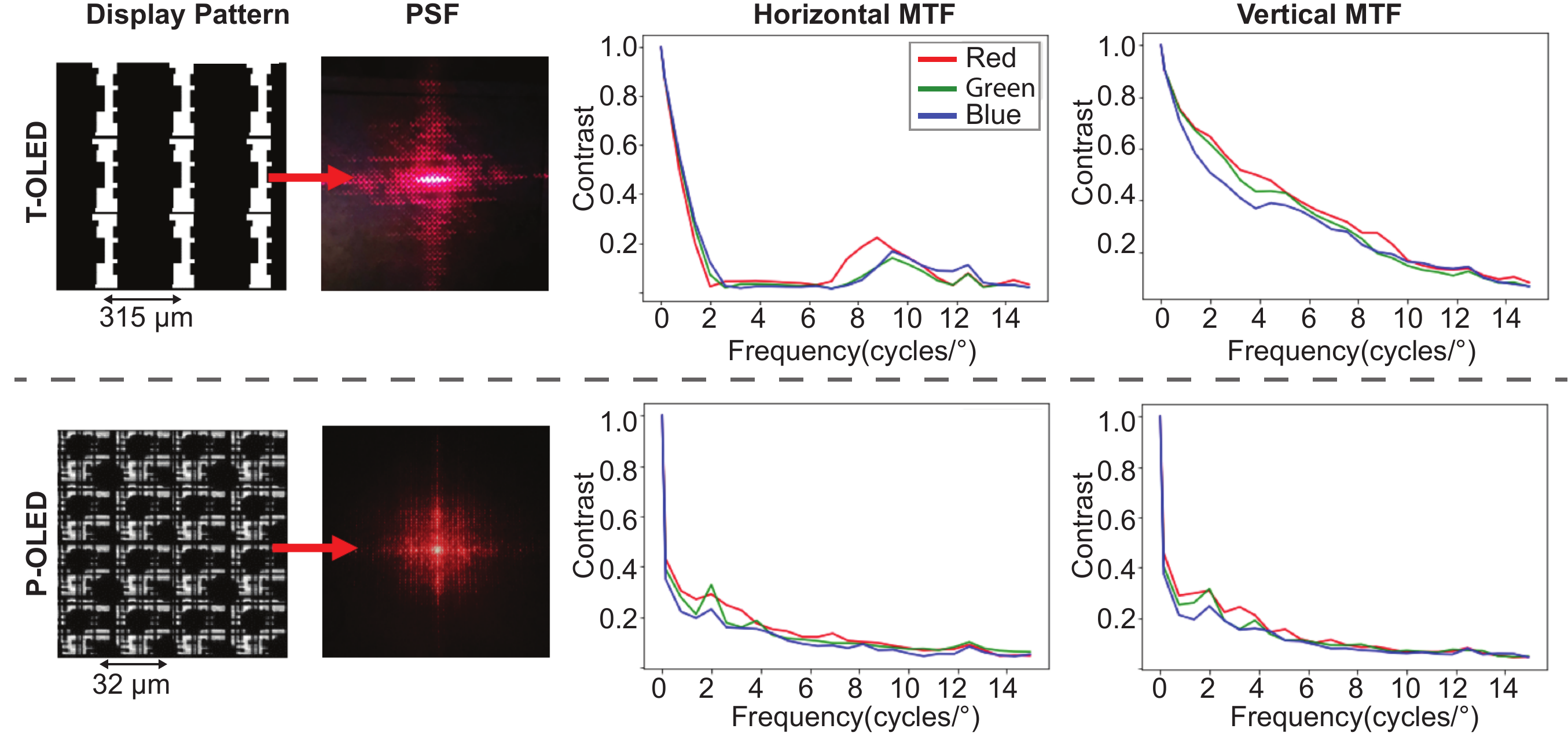}
		\caption{} 
	\end{subfigure}
	\caption{Image formation pipeline of under-display camera (UDC) problem. (a) Image Formation Pipeline. (b)Optics characteristics of UDC. The structure of the 4K T-OLED has a grating-like pixel layout. P-OLED differs from T-OLED in sub-pixel design. 
	From left to right: Micrography of display patterns, PSFs (red light only) and MTFs (red, green, and blue).}
	\label{fig:os}
	\vspace{-4mm}
\end{figure*}

\section{Formulation}
In this section, we discuss the optical system and image formation process of the proposed UDC imaging system. We analyze the degradation type, light transmission rate and visualize the Point Spread Function (PSF). Moreover, we formulate the image formation pipeline to compute simulated PSF from measurements. 
\subsection{Optical System Analysis}\label{sec:opt}
\textbf{Optical Elements.}
In our experiments, we focus on the Organic Light-Emitting Diode (OLED) displays~\cite{wenke2016organic} as they have superior optical properties compared to the traditional LCDs (Liquid Crystal Display). Due to confidentiality reasons it is often difficult to obtain the sample materials used for demos from commercial companies. In this case, we select the displays with different transparencies to improve the generalization. Note that all the displays are \textbf{non-active} in our experiments, since in real scenario, the display can be turned off locally by setting black pixels on local regions of the OLED display when the camera is in operation to 1) reduce unnecessary difficulty from display contents while not affecting user experience and 2) provide users with the status of the device and thus ensure privacy. 

Owing to transparent materials being used in OLED display panels, visible lights can be better transmitted through the OLEDs than LCDs. In the meantime, pixels are also arranged such that open area is maximized. In particular, we focus on 4k Transparent OLED (T-OLED) and a phone Pentile OLED (P-OLED). Figure \ref{fig:os} is a micrograph illustration of the pixel layout in the two types of OLED displays. The structure of the 4K T-OLED has a grating-like pixel layout. P-OLED differs from T-OLED in sub-pixel design. It follows the basic structure of RGBG matrix. 

\begin{table}[t]
\setlength{\abovecaptionskip}{0pt}
\centering
\footnotesize
\caption{Comparison of two displays in terms of light transmission rate and physical pixel layout and open areas.}
\resizebox{\columnwidth}{!}{
\begin{tabular}{|l|c|c|}
\hline
{Metrics}&{T-OLED}&P-OLED\\ \hline \hline
Pixel Layout Type&Stripe&Pentile \\
Open Area&21$\%$ &23$\%$ \\
Transmission Rate &20$\%$ &2.9$\%$\\
Major Degradation &Blur, Noise&Low-light, Color Shift, Noise \\
\hline
\end{tabular}}
\label{exp:measure}
\vspace{-4mm}
\end{table}

\textbf{Light Transmission Rate.}
We measure the transmission efficiency of the OLEDs by using a spectrophotometer and white light source. Table \ref{exp:measure} compares the light transmission rate of the two displays. For T-OLED, the open area occupies about 21$\%$, and the light transmission rate is around 20$\%$. For P-OLED, although the open area can be as large as 23$\%$, the light transmission rate is only 2.9$\%$. 

The loss of photons can be attributed mainly to the structure of P-OLED. First, the P-OLED has a finer pixel pitch, so photos are scattered to higher angles comparing to the T-OLED. As a result, high angle photons are not collected by the lens. Second, P-OLED is a flexible/bendable display, which has a poly-amide substrate on which the OLED is formed. Such a substrate has relatively low transmission efficiency, causing photons to be absorbed. The absorbed light with certain wavelengths may make the images captured through a polyamide-containing display panel by a UDC appear yellow. 
As a result, imaging through a P-OLED results in lower signal-to-noise ratio (SNR) comparing to using a T-OLED, and has a color shift issue. One real imaging example is shown in Figure \ref{fig:realsample}.

\textbf{Diffraction Pattern and Point Spread Function (PSF).}
Light diffracts as it propagates through obstacles with sizes that are similar to its wavelength. Unfortunately, the size of the openings in the pixel layout is on the order of wavelength of visible light, and images formed will be degraded due to diffraction. 

Here we characterize our system by measuring the point spread function (PSF). We do so by pointing a collimated red laser beam  ($\lambda =$ 650nm) at the display panel and recording the image formed on the sensor, as demonstrated in Figure \ref{fig:schematic_3d} and \ref{fig:os}. An ideal PSF shall resemble a delta function, which then forms a perfect image of the scene. However, light greatly spreads out in UDC. For T-OLED, light spreads mostly across the horizontal direction due to its nearly one dimensional structure in the pixel layout, while for P-OLED, light is more equally distributed as the pixel layout is complex. Therefore, images captured by UDC are either blurry (T-OLED) or hazy (P-OLED).

\textbf{Modulation Transfer Function (MTF)}
Modulation Transfer Function (MTF) is another important metric for an imaging system, as it considers the effect of finite lens aperture, lens performance, finite pixel size, noise, non-linearities, quantization (spatial and bit depth), and diffraction in our systems. We characterize the MTF of our systems by recording sinusoidal patterns with increasing frequency in both lateral dimensions, and we report them in Figure ~\ref{fig:os}. For T-OLED, contrasts along the horizontal direction are mostly lost in the mid-band frequency due to diffraction. This phenomenon is due to the nearly one-dimensional pixel layout of the T-OLED. Figure ~\ref{fig:realsample} shows severe smearing horizontally when putting T-OLED in front of the camera. While for P-OLED, the MTF is almost identical to that of display-free camera, except with severe contrast loss. Fortunately, however, nulls have not been observed in any particular frequencies.

\subsection{Image Formation Pipeline}
In this section, we derive the image formation process of UDC based on the analysis in the previous sections. Given a calibrated pixel layout and measurements using a specific camera, degraded images can be simulated from a scene. From the forward model, we can compute the ideal PSF and consequently synthesize datasets from ground truth images. 

Given an object in the scene $\mathbf{x}$, the degraded observation $\mathbf{y}$ can be modeled by a convolution process,
\begin{equation}
	\mathbf{y} = (\gamma \mathbf{x}) \otimes \mathbf{k}+ \mathbf{n},
\label{eq:degradation}
 \end{equation}
where $\gamma$ is the intensity scaling factor under the current gain setting and display type, $\mathbf{k}$ is the PSF, and $\mathbf{n}$ is the zero-mean signal-dependent noise. Notice that this is a simple noise model that approximately resembles the combination of shot noise and readout noise of the camera sensor, and it will be discussed in a later section.

\textbf{Intensity Scaling Factor ($\gamma$)}
The intensity scaling factor measures the changing ratio of the average pixel values after covering the camera with a display. It simultaneously relates to the physical light transmission rate of the display, as well as the digital gain $\delta$ setting of the camera. $\gamma$ can be computed from the ratio of $\delta$-gain amplified average intensity values $I_d(\delta, s)$ at position $s$ captured by UDC, to the 0-gain average intensity values $I_{nd}(0, s)$ by naked camera within an enclosed region $S$. It is represented by,
\begin{equation}
	\gamma = \frac{\int_S I_d(\delta, s)ds}{\int_S I_{nd}(0, s)ds}
\label{eq:gamma}
 \end{equation}
 
\textbf{Diffraction Model}
We approximate the blur kernel $\mathbf{k}$, which is the Point Spread Function (PSF) of the UDC. As shown in Figure \ref{fig:schematic_3d}, in our model, we assume the display panel is at the principle plane of the lens. We also assume the input light is monochromatic plane wave with wavelength $\lambda$ (i.e. perfectly coherent), or equivalently light from a distance object with unit amplitude. Let the display pattern represented by transparency with complex amplitude transmittance be $g(m, n)$ at the Cartesian co-ordinate $(m,n)$, and let the camera aperture/pupil function $p(m,n)$ be 1 if $(m, n)$ lies inside the lens aperture region and 0 otherwise, then the display pattern inside the aperture range $g_p(m,n)$ becomes,
\begin{equation}
    g_p(m,n) = g(m,n)p(m,n).
   \label{eq:crop}
\end{equation}

At the focal plane of the lens (i.e. 1 focal length away from the principle plane), the image measured is the intensity distribution of the complex field, which is proportional to the Fourier transform of the electric field at the principle plane~\cite{goodman2005introduction}:


\begin{equation}
    I(u,v) \propto 
    \left\vert {\iint}^{\infty}_{-\infty}g_p(m,n)\exp \left[-j\frac{2\pi}{\lambda f}(mu+nv)\right] \text{d}m\text{d}n\right\vert^2. 
\end{equation}
Suppose $G_p(v_m, v_n) = \mathscr{F}(g_p(m,n))$, where $\mathscr{F}(\cdot)$ is the Fourier transform operator, then
\begin{equation}
I(u,v) \propto \left\vert G_p(v_m, v_n)\right\vert^2 = \left\vert G_p(\frac{u}{\lambda f}, \frac{v}{\lambda f})\right\vert^2,
\end{equation}
which performs proper scaling on the Fourier transform of the display pattern on the focal plane.

Therefore, to compute the PSF $\mathbf{k}$ for image $\mathbf{x}$, we start from computing Discrete Fourier Transform (DFT) with squared magnitude $M(a,b) = |\hat{G_p}(a,b)|^2$ of the $N \times N$ microscope transmission images $\hat{g_p}$ of the display pattern and re-scaling it. Then, the spatial down-sampling factor $r$ (denoted by $\downarrow r$) becomes,
\begin{equation}
    r = \frac{1}{\lambda f} \cdot {\delta_N N} \cdot {\rho},
    \label{eq:ratio}
\end{equation}
where $\delta_N$ is the pixel size of the $\hat{g_p}$ images, and $\rho$ is the pixel size of the sensor. Finally, $\mathbf{k}$ can be represented as
\begin{equation}
    k(i,j) = \frac{M_{\downarrow r} (i,j)}{\sum_{(\hat{i}, \hat{j})}M_{\downarrow r}(\hat{i}, \hat{j})}.
\end{equation}
$k$ is a normalized form since we want to guarantee that it represents the density distribution of the intensity with diffraction effect. Note that only PSF for a single wavelength is computed for simplicity. However, scenes in the real-world are by no means monochromatic. Therefore, in order to calculate an accurate color image from such UDC systems, PSF for multiple wavelengths shall be computed. More details will be shown in Section \ref{sec:data}.

\textbf{Adding Noises}
We follow the commonly used shot-read noise model \cite{brooks2019unprocessing,hasinoff2014photon,liu2007automatic} to represent the real noise on the imaging sensor. Given the dark and blur signal $w = (\gamma \mathbf{x}) \otimes \mathbf{k}$, the shot and readout noise can be modeled by a heteroscedastic Gaussian,
\begin{equation}
	\mathbf{n} \sim \mathcal{N}(\mu=0, \sigma^2 = \lambda_{read} + \lambda_{shot} w), 
\label{eq:conv}
 \end{equation}
where the variance $\sigma$ is signal-dependent, and $\lambda_{read}$ , $\lambda_{shot}$ are determined by camera sensor and gain values. 

\section{Data Acquisition and Synthesis}
We propose an image acquisition system called Monitor-Camera Imaging System (MCIS). In particular, we display natural images with rich textures on high-resolution monitor and capture them with a static camera. The method is more controllable, efficient, and automatic to capture a variety of scene contents than using mobile set-ups to capture limited static objects or real scenes. 
\subsection{Monitor-Camera Imaging System}
\begin{figure}[t]
	\centering
		\includegraphics[width=\linewidth]{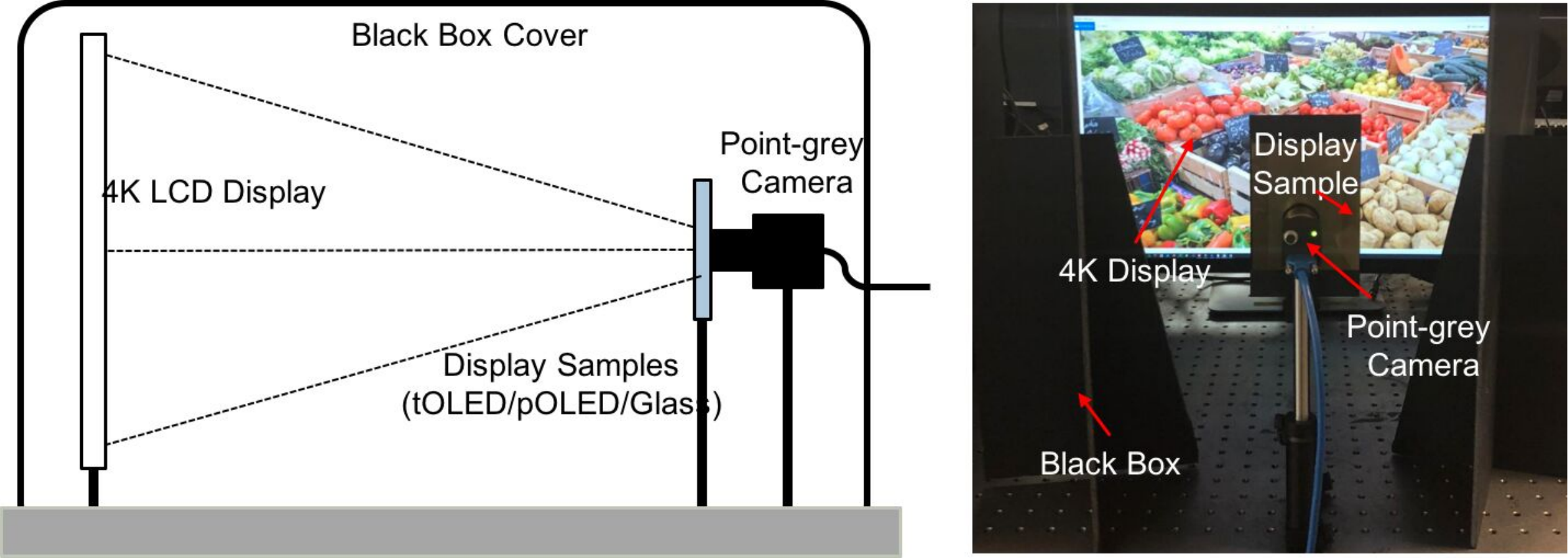}
	\caption{Monitor-Camera Imaging System (MCIS). MCIS consists of a 4K LCD monitor, the 2K FLIR RGB Point-Grey research camera, and a panel that is either T-OLED, P-OLED or Glass(i.e. no display). The camera is mounted on the center line of the 4K monitor, and adjusted to cover the full monitor range.}
	\label{fig:dcis}
\end{figure}
\begin{figure}[t]
	\centering
	\begin{subfigure}{0.32\columnwidth} 
		\includegraphics[width=\columnwidth]{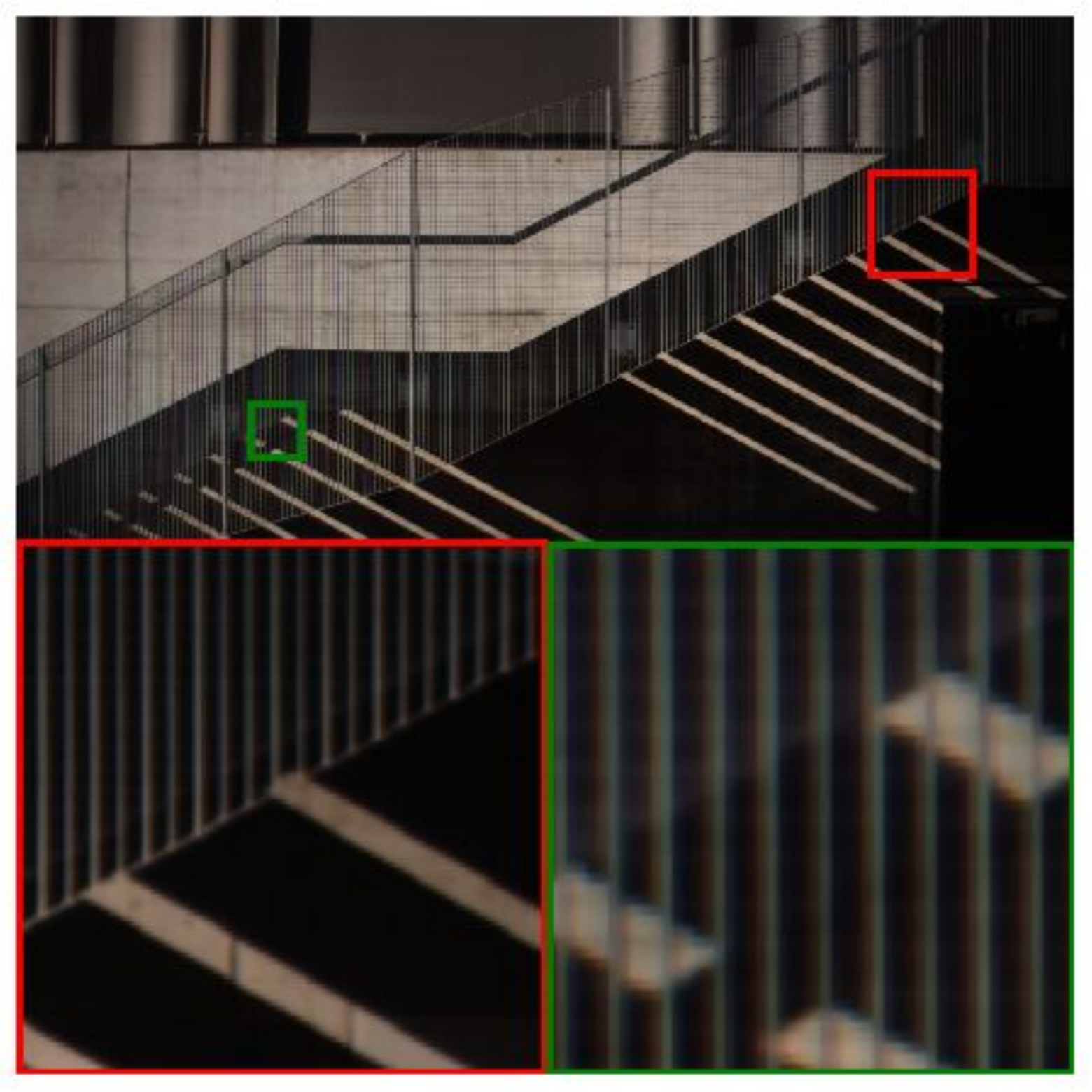}
		\caption{Display-free}
	\end{subfigure}
	\begin{subfigure}{0.32\columnwidth} 
		\includegraphics[width=\columnwidth]{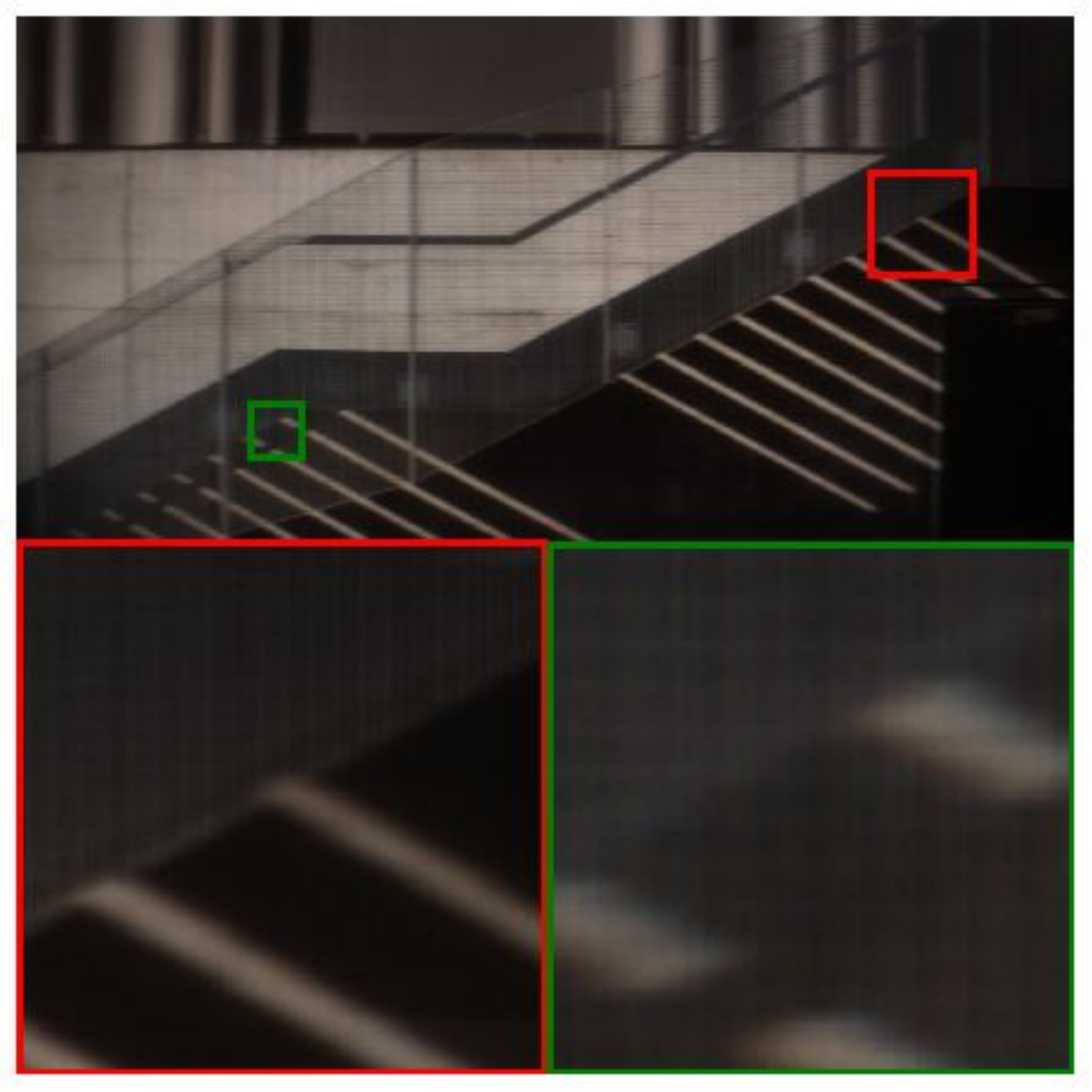}
		\caption{TOLED}
	\end{subfigure}
    \begin{subfigure}{0.32\columnwidth} 
		\includegraphics[width=\columnwidth]{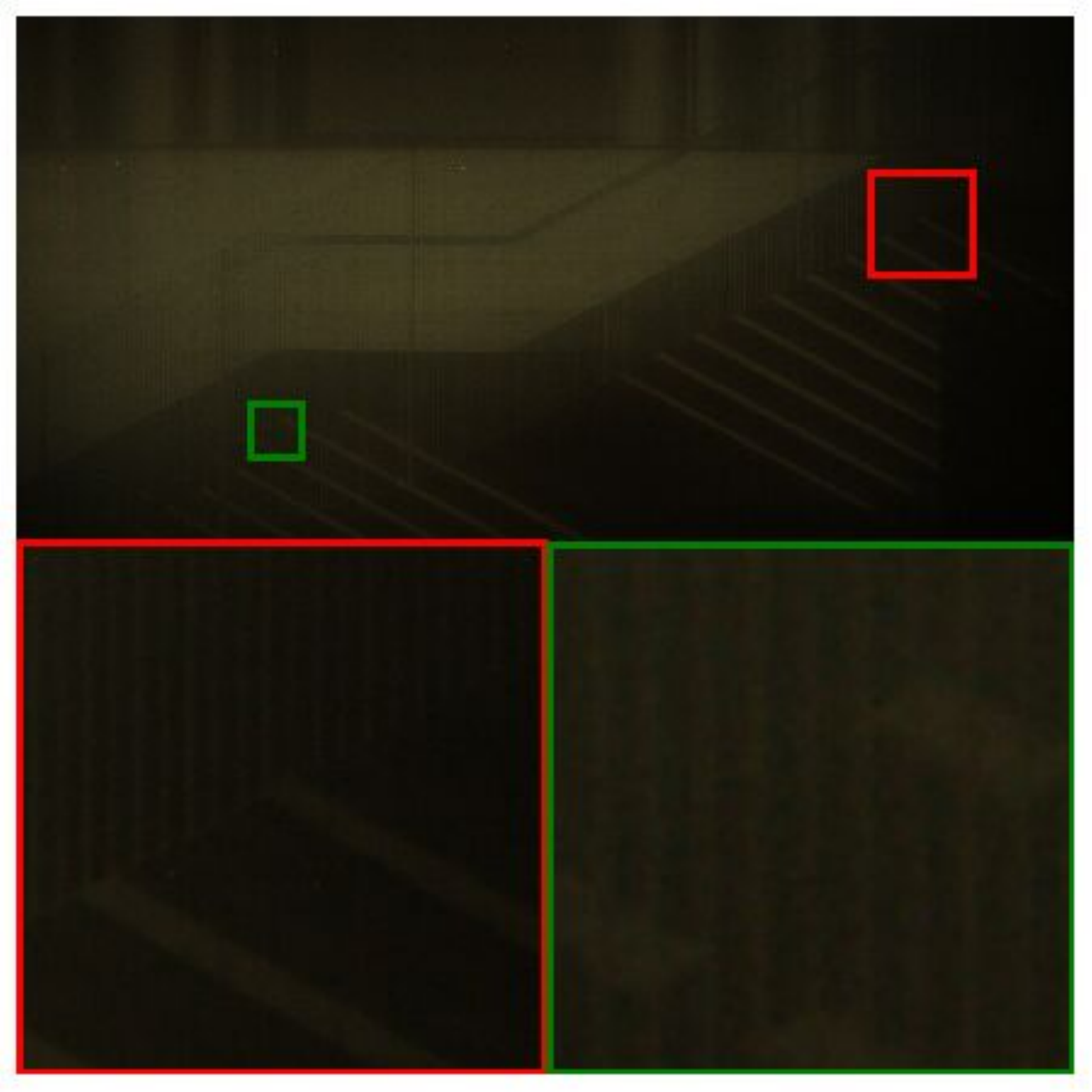}
		\caption{POLED}
	\end{subfigure}
	\caption{Real samples collected by the proposed MCIS. Images captured by T-OLED are blur and noisy, while those captured by P-OLED are low-light, color-shifted and hazy. }
	\label{fig:realsample}
	\vspace{-4mm}
\end{figure}
The system architecture is shown in Figure \ref{fig:dcis}. MCIS consists of a 4K LCD monitor, the 2K FLIR RGB Point-Grey research camera, and a panel that is either T-OLED, P-OLED or Glass(i.e. no display). The camera is mounted on the center line of the 4K monitor, and adjusted to cover the full monitor range. We calibrate the camera gain by measuring a $256 \times 256$ white square shown on the monitor and matching the RGB histogram. For fair comparison and simplicity, we adjust the focus and fix the aperture to f/1.8. It guarantees a reasonable pixel intensity range avoiding saturation while collecting data with no gain. Suppose we develop a real-time video system, the frame rate has to be higher than 8 fps. So the lowest shutter speed is 125 ms for the better image quality and the higher Signal-to-Noise Ratio (SNR). 

\begin{table}[t]\setlength{\tabcolsep}{12pt}
\setlength{\abovecaptionskip}{0pt}
\centering
\footnotesize
\caption{Camera Settings for different set of collected data}
\resizebox{\columnwidth}{!}{
\begin{tabular}{|c|c|c|c|}
\hline
{Parameteres}&{No-Display}&T-OLED&P-OLED\\ \hline \hline 
Aperture&\multicolumn{3}{c|}{f/1.8} \\\hline 
FPS/Shutter&\multicolumn{3}{c|}{8/125ms}  \\\hline 
Brightness&\multicolumn{3}{c|}{0} \\\hline 
Gamma&\multicolumn{3}{c|}{1} \\\hline 
Gain&1&6&25(Full)\\\hline 
White-balance&Yes&None&None\\
\hline
\end{tabular}}
\label{tab:campara}
\vspace{-4mm}
\end{table}

We split 300 images from DIV2K dataset \cite{agustsson2017ntire}, and take turns displaying them on a 4K LCD in full screen mode. We either rotate or resize the images to maintain the Aspect Ratio. For training purposes, we capture two sets of images, which are the degraded images $\{y_i\}$, and the degradation-free set $\{x_i\}$.

To capture $\{x_i\}$, we first cover the camera with a thin glass panel which has the same thickness as a display panel. This process allows us to avoid the pixel misalignment issues caused by light refraction inside the panel. To eliminate the image noises in $\{x_i\}$, we average the 16 repeated captured frames. Then we replace the glass with a display panel (T-OLED or P-OLED), calibrate the specific gain value avoiding saturation, and capture $\{y_i\}$. For each set, we record both the 16-bit 1-channel linear RAW CMOS sensor data as well as the 8-bit 3-channel linear RGB data after in-camera pipeline that includes demosaicing. The collected pairs are naturally well spatially-aligned in pixel-level. They can be directly used for deep model training without further transformations.

Due to the yellow substrate inside the P-OLED, certain light colors, especially blue, are filtered out and changes the white balance significantly. We therefore did not further alter the white balance. The light transmission ratio of P-OLED is extremely low, so we set up the gain value to be the maximum (25) for higher signal values. All the detailed camera settings for the two display types are shown in Table \ref{tab:campara}. One real data sample is shown in Figure \ref{fig:realsample}. As discussed and analyzed in Section \ref{sec:opt}, images captured by T-OLED are blur and noisy, while those captured by P-OLED are low-light, color-shifted and hazy.

\begin{table}[t]\setlength{\tabcolsep}{8pt}
\setlength{\abovecaptionskip}{0pt}
\centering
\footnotesize
\caption{Measured parameters for data synthesis}
\resizebox{\columnwidth}{!}{
\begin{tabular}{|l|c|c|c|c|c|c|}
\hline
{Parameteres}&\multicolumn{3}{c|}{T-OLED}&\multicolumn{3}{c|}{P-OLED}\\ \hline\hline 
&R&G&B&R&G&B\\\hline 
$\gamma$       &0.97 &0.97 &0.97 &0.34 &0.34 &0.20\\\hline 
$\lambda$  (nm)&640 &520 &450 &640 &520 &450\\\hline 
r&2.41&2.98&3.44&2.41&2.98&3.44\\\hline 
\end{tabular}}
\label{tab:factor}
\vspace{-7mm}
\end{table}

\subsection{Realistic Data Synthesis Pipeline}\label{sec:data}
We follow the image formation pipeline to simulate the degradation on the collected $\{x_i\}$. A model-based data synthesis method will benefit concept understanding and further generalization. Note that all the camera settings are the same as the one while collecting real data. 
We first transform the 16-bit raw sensor data $\{x_i\}$ into four bayer channels $x_r$, $x_{gr}$, $x_{gl}$, and $x_b$. Then, we multiply the measured intensity scaling factor $\gamma$, compute the normalized and scaled PSF $k$, and add noises to the synthesize degraded data.

\textbf{Measuring $\gamma$}: To measure $\gamma$ for each channel using the MCIS, we select the region of interest $S$ to be a square region of size $256 \times 256$, and display the intensity value input from 0 to 255 with stride 10 on the monitor. We then record the average intensity both with and without the display for each discrete intensity value, and plot the relationship between display-covered values and no-display-covered ones. Using linear regression, we obtain the ratios of lines for different RGGB channel. For T-OLED, the measured $\gamma$ is 0.97, same for all the channels. For P-OLED, $\gamma=0.20$ for the blue channel, and $\gamma=0.34$ for the other three channels. 

\textbf{Computing PSF}: Following equation \ref{eq:crop}, we acquire the transmission microscope images of the display pattern and crop them with the approximated circular aperture shape with diameter $3333\mu m$, the size of the camera aperture. In equation \ref{eq:ratio}, the $\delta_N N$ is $3333\mu m$. $\rho$ equals to $1.55 \mu m/pixel$ in Sony sensor. However, after re-arranging the raw image into four RGGB channels, $\rho$ becomes 3.1 for each channel. The focal length is $6000 \mu m$. $\lambda=(640, 520, 450)$ for R, G, B channel, which are the approximated center peaks of the R, G, B filters respectively on the sensor. It yields the down-sampling ratio $r=(2.41, 2.98, 3.44)$ for the R, G, and B channels.

\textbf{Adding Noises}: We measure $\lambda_{read}$ and $\lambda_{shot}$ to estimate the noise statistics. We display random patterns within the $256 \times 256$ window on the monitor, collect the paired noisy and noise-free RAW data, and compute their differences. For each of the RGGB channel, we linearly regress the function of noise variance to the intensity value, and obtain the ratio as the shot noise variance, and the y-intersection as the readout noise variance. We then repeat the process 100 times and collect pairs of data points. Finally, we estimate the distribution and randomly sample $\lambda_{read}$ and $\lambda_{shot}$. All the measurements are listed in Table \ref{tab:factor}.
\begin{figure}[t]
	\centering
		\includegraphics[width=\columnwidth]{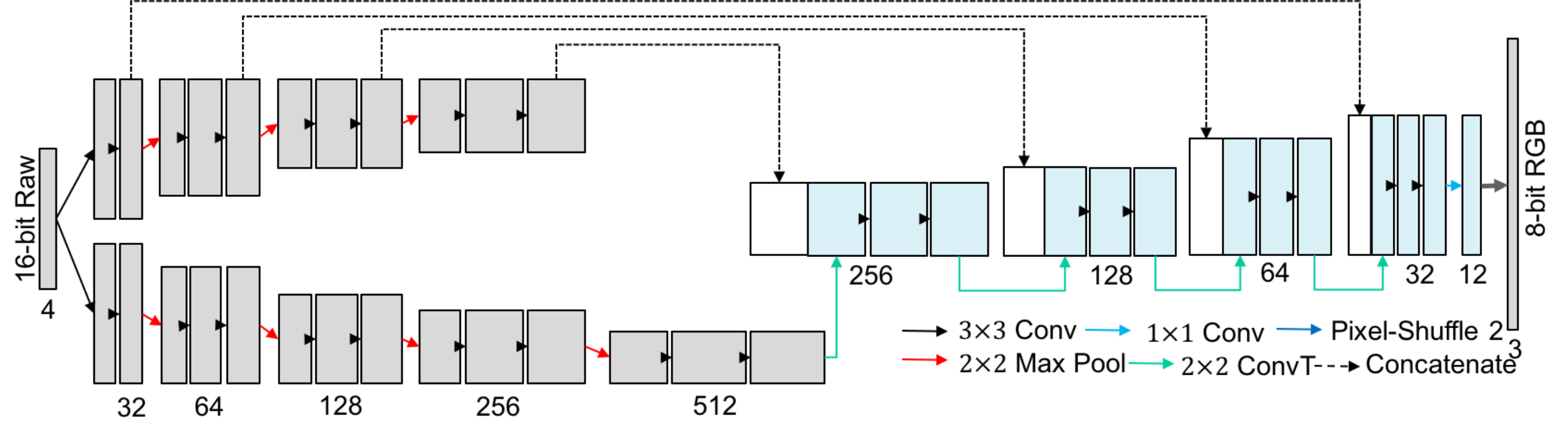}
	\caption{Network structure of the proposed UNet. It takes a 4-channel RAW sensor data observation $y$, and outputs the restored 3-channel RGB image $x$. }
	\label{fig:structure}
	\vspace{-6mm}
\end{figure}

\section{Image Restoration Baselines}
We use the collected real paired data, synthetic paired data, simulated PSF, and all the necessary measurements to perform image restoration. We split the 300 pairs of images in the UDC dataset into 200 for training, 40 for validation and 60 images in the testing partition. All the images have a resolution of $1024 \times 2048$.

\subsection{Deconvolution Pipeline (DeP)} The DeP is a general-purpose conventional pipeline concatenating denoising and deconvolution (Wiener Filter), which is an inverse process of the analyzed image formation. To better utilize the unsupervised Wiener Filter (WF) \cite{orieux2010bayesian}, we first apply the BM3D denoiser to each RAW channel separately, afterwards we linearly divide the measured $\gamma$ with the outputs for intensity scaling. After that, WF is applied to each channel given the pre-computed PSF $\mathbf{k}$. Finally, RAW images with bayer pattern are demosaiced by linear interpolation. The restored results are evaluated on the testing partition of the UDC dataset.

\subsection{Learning-based Methods}
\textbf{UNet.} We propose a learning-based restoration network baseline as shown in Figure \ref{fig:structure}. The proposed model takes a 4-channel RAW sensor data observation $y$, and outputs the restored 3-channel RGB image $x$. The model conducts denoising, debluring, white-balancing, intensity scaling, and demosaicing in a single network, whose structure is basically a UNet. We split the encoder into two sub-encoders, one of which is for computing residual details to add, and the other one learns content encoding from degraded images. By splitting the encoder, compared with doubling the width of each layer, we will have fewer parameters, and make the inference and learning more efficient. To train the model from paired images, we apply the $L_1$ loss, which will at large guarantee the temporal stability compared with adversarial loss \cite{goodfellow2014generative}. Besides, we also apply $SSIM$ and Perception Loss (VGG Loss) for ablation study. 

We crop patches of $256 \times 256$, and augment the training data using the raw image augmentation \cite{liu2019learning} while preserving the RGGB bayer pattern. We train the model for 400 epochs using Adam optimizer ($\beta_1=0.9$, $\beta_2=0.999$ and $\epsilon=10^{-8}$) with learning rate $10^{-4}$ and decay factor 0.5 after 200 epoches. We also train the same structure using the synthetic data (denoted as \textbf{UNet(Syn)}) generated by the pipeline proposed in section \ref{sec:data}.

\textbf{ResNet.} Additionally, a data-driven ResNet trained with the same data is utilized for evaluation. To our knowledge, UNet and ResNet-based structures are two widely-used deep models for image restoration. We use 16 residual blocks with a feature width of 64 for our ResNet architecture, just as Lim \textit{et al.} do for EDSR \cite{lim2017enhanced}. The model also takes 4-channel RAW data, and outputs 3-channel RGB images. The data-driven models cannot be directly adaptive to UDC inputs if only trained with bi-cubic degradation. We did not compare with their model structures because model novelty is not our main claim, and the presented two methods are the most general ways which can achieve real-time inference as the baselines. Other model variants can be further explored in future work. 

\begin{figure*}[t]\setlength{\belowcaptionskip}{-3pt}
\setlength{\abovecaptionskip}{3pt}
	\centering
	\begin{subfigure}{0.19\linewidth} 
		\includegraphics[width=\textwidth]{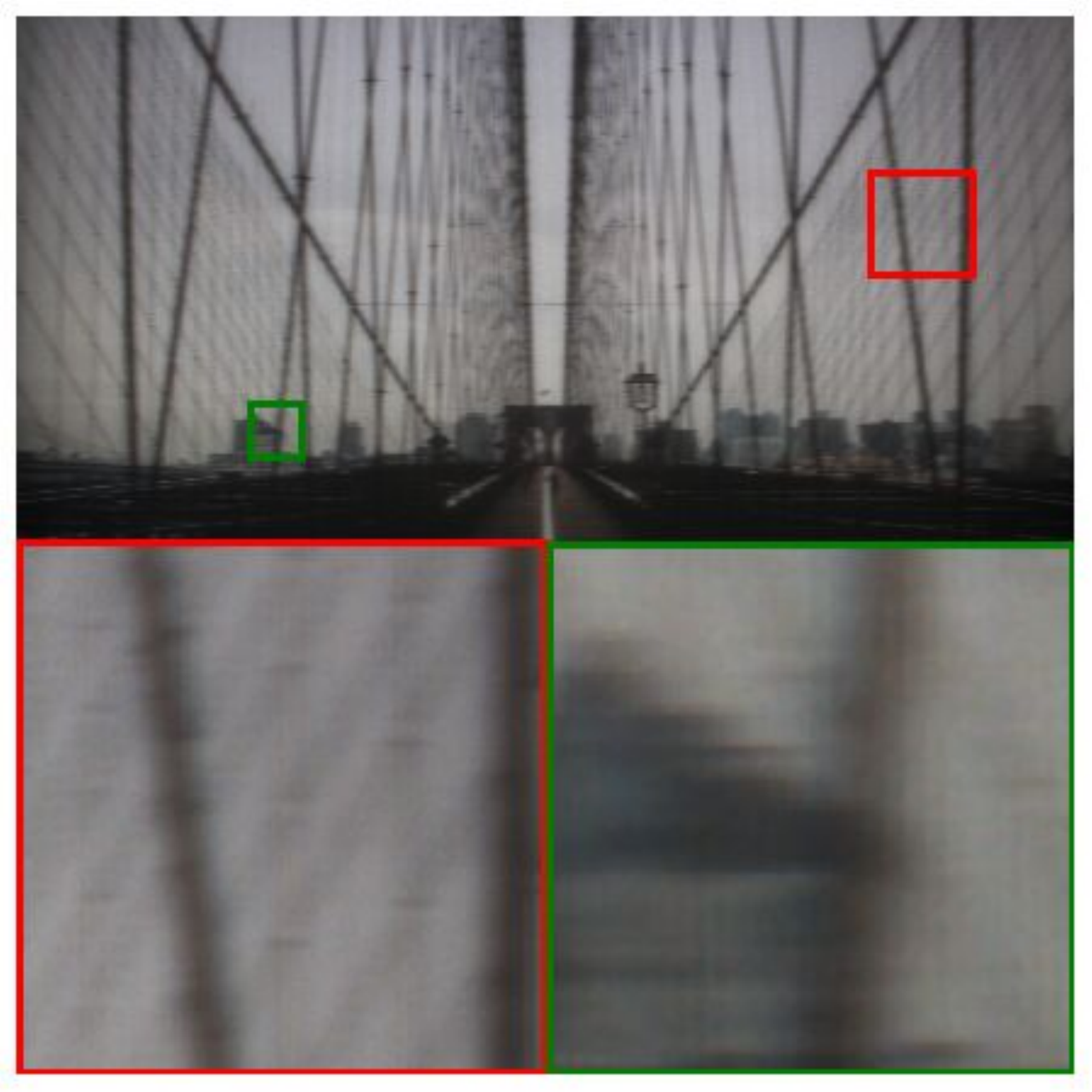}
	\end{subfigure}
	\begin{subfigure}{0.19\linewidth} 
		\includegraphics[width=\textwidth]{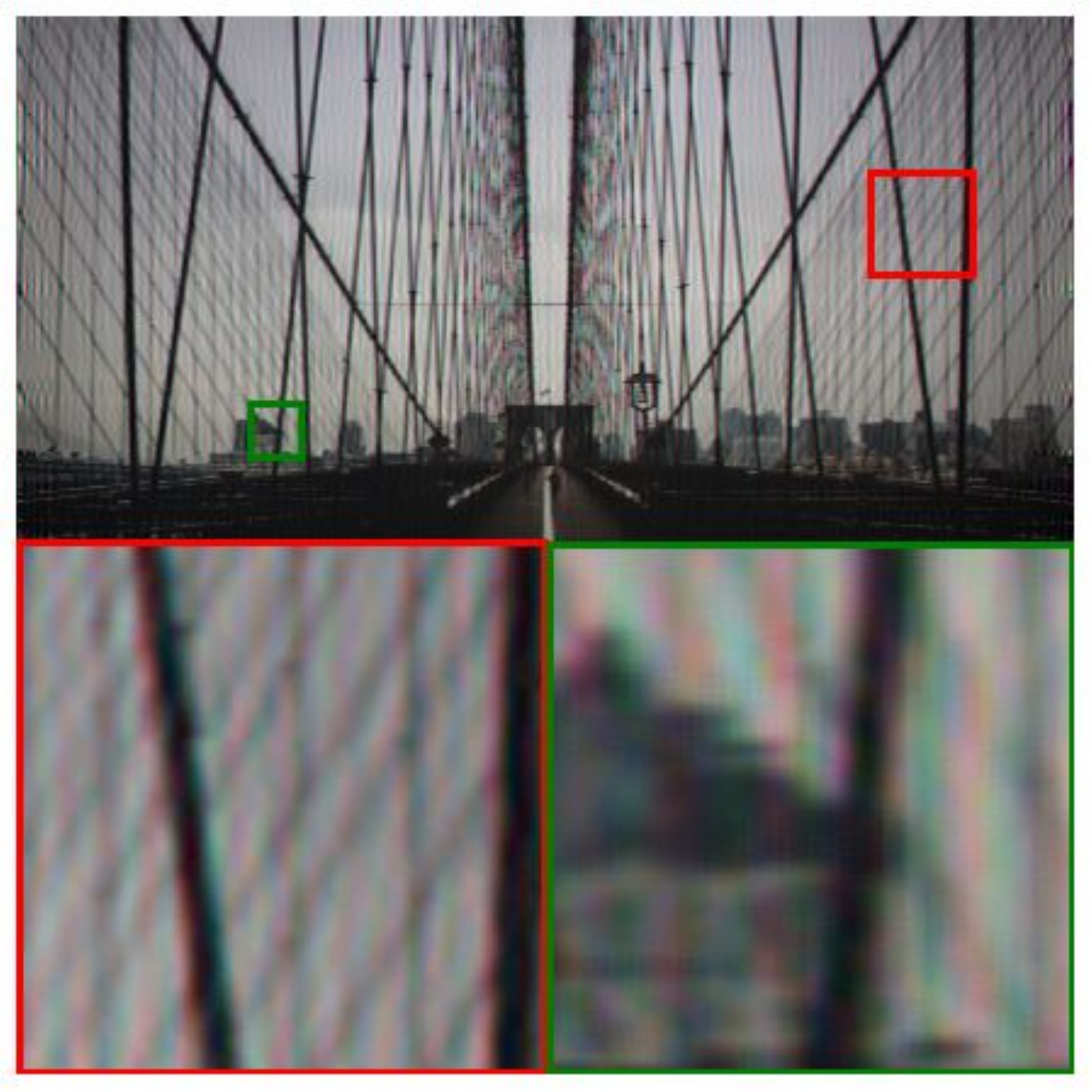}
	\end{subfigure}
		\begin{subfigure}{0.19\linewidth} 
		\includegraphics[width=\textwidth]{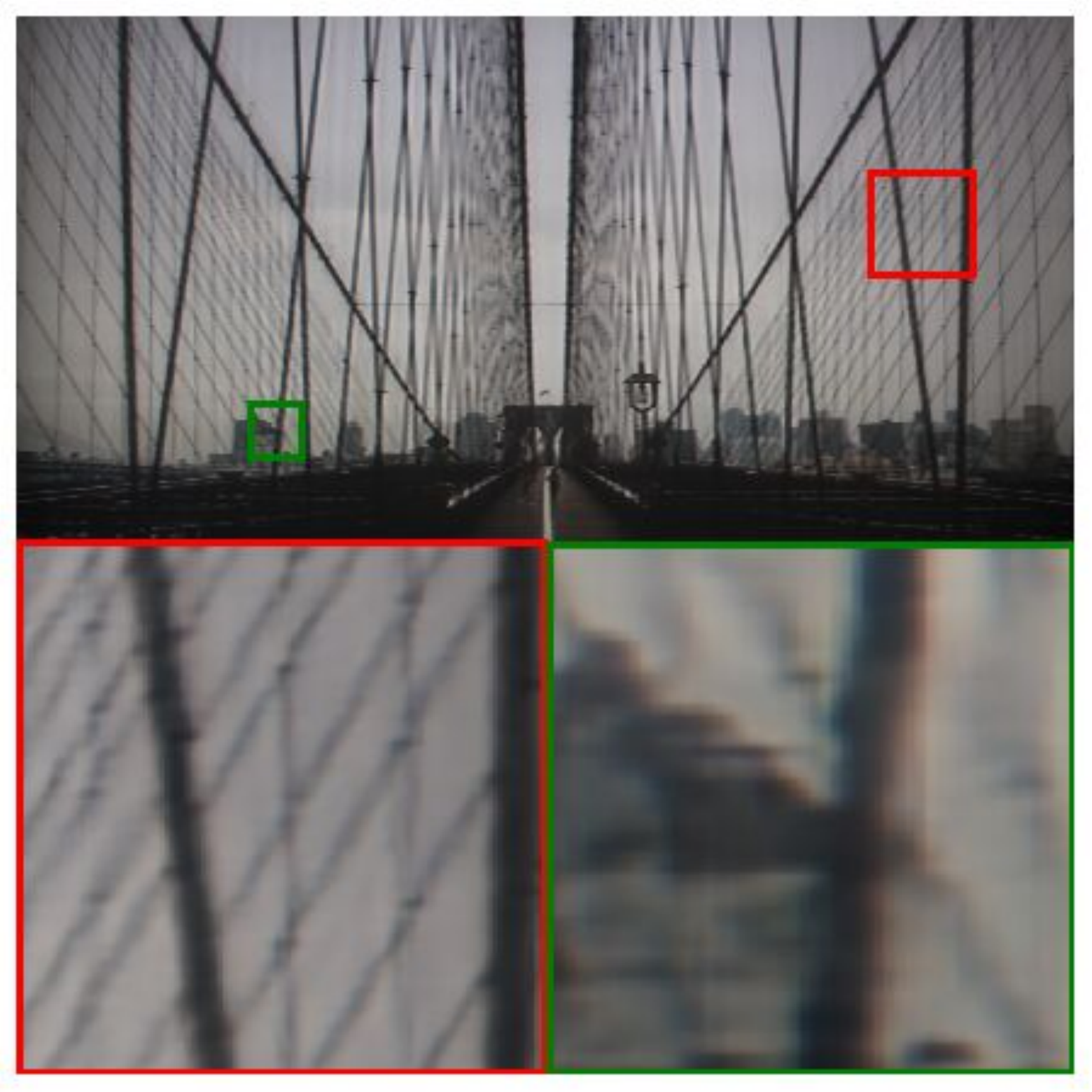}
	\end{subfigure}
	\begin{subfigure}{0.19\linewidth} 
		\includegraphics[width=\textwidth]{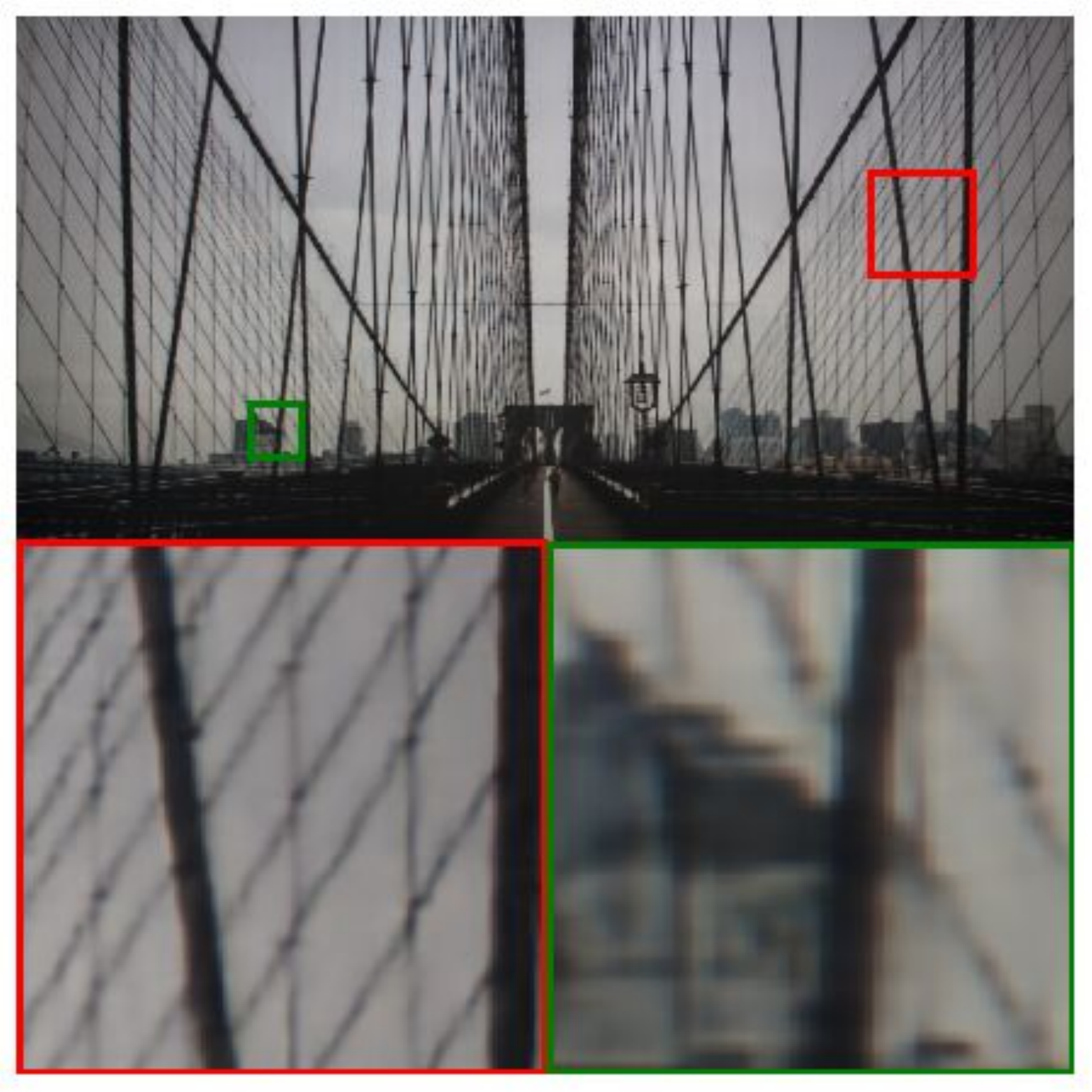}
	\end{subfigure}
	\begin{subfigure}{0.19\linewidth} 
		\includegraphics[width=\textwidth]{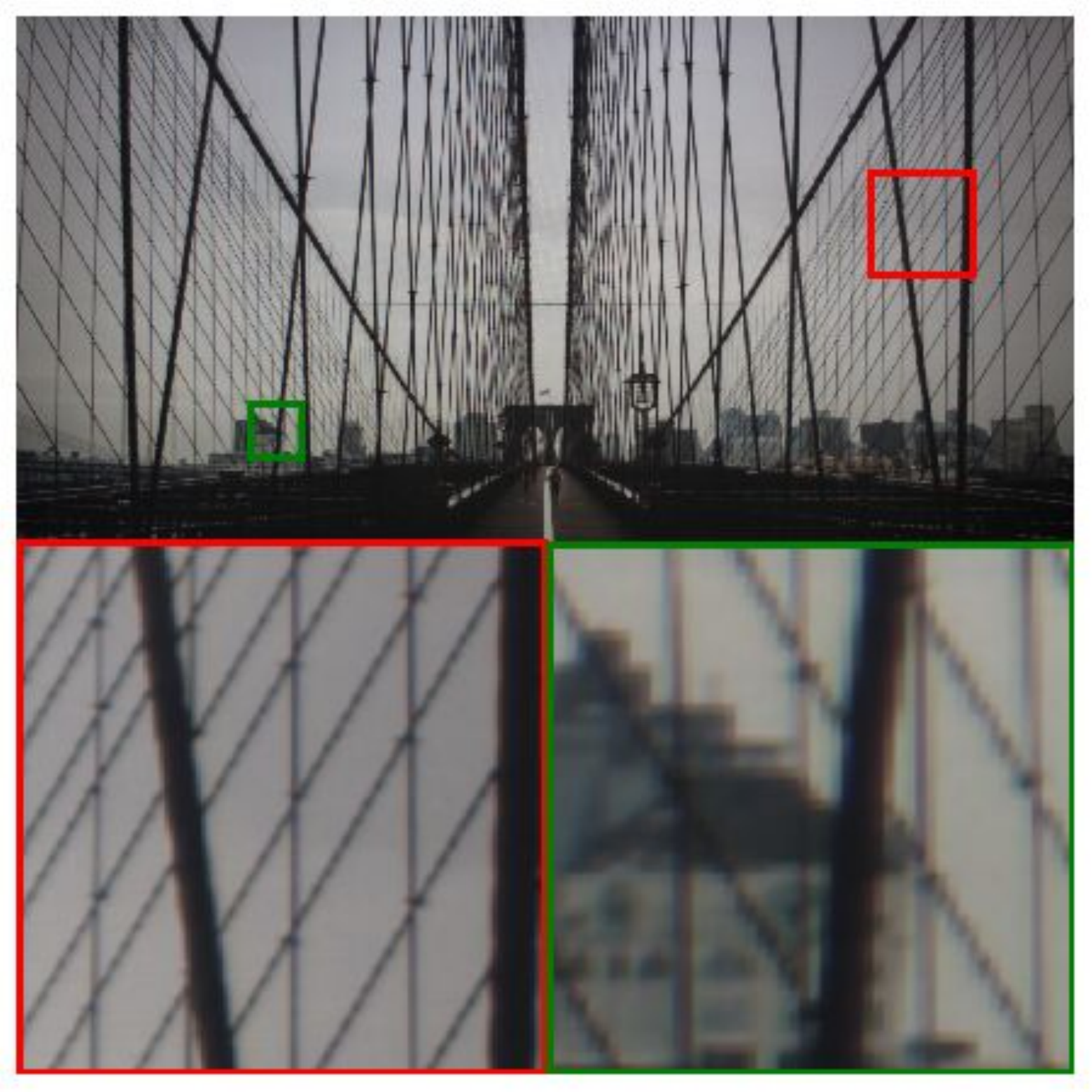}
	\end{subfigure}
	\begin{subfigure}{0.19\linewidth} 
		\includegraphics[width=\textwidth]{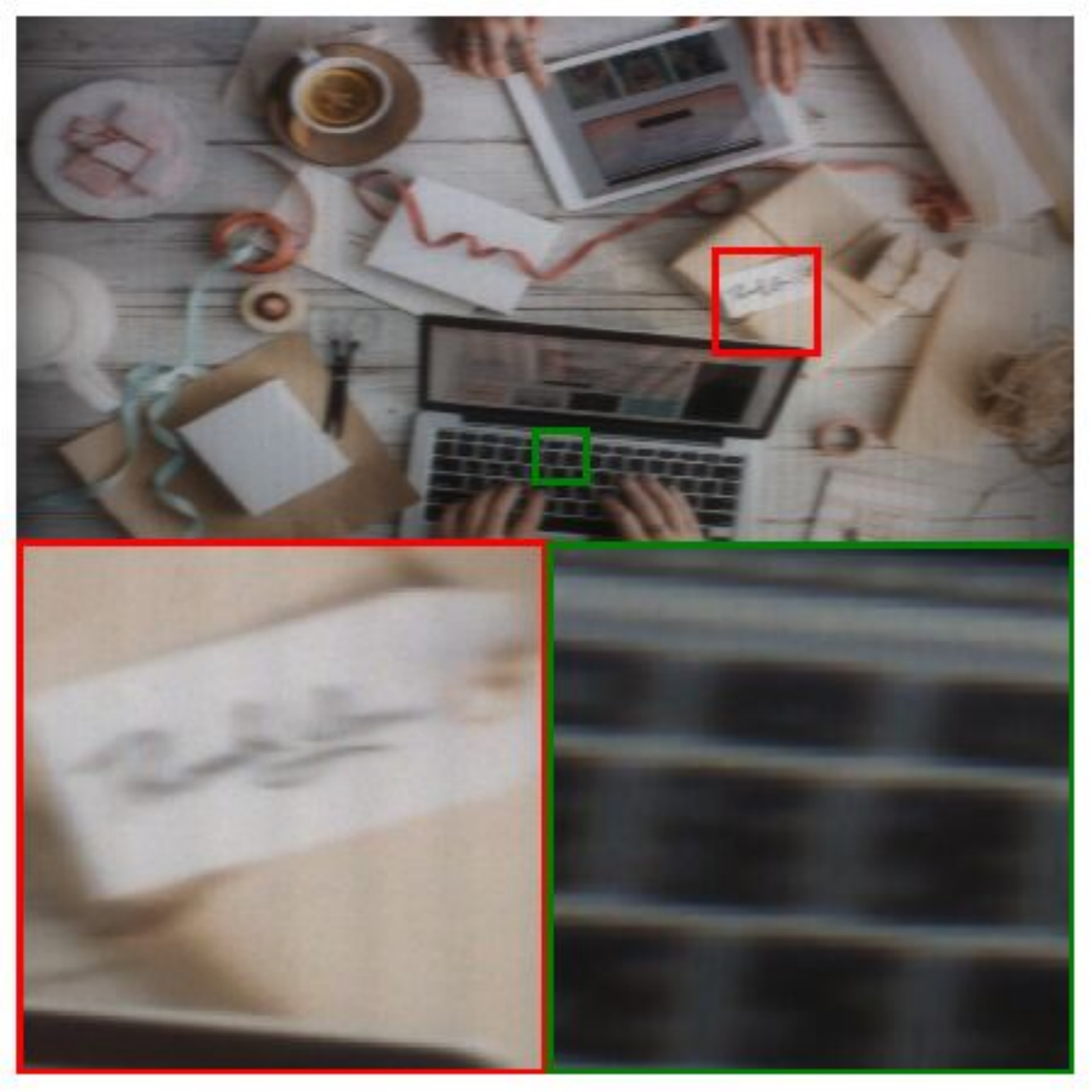}
		\caption{T-OLED} 
	\end{subfigure}
	\begin{subfigure}{0.19\linewidth} 
		\includegraphics[width=\textwidth]{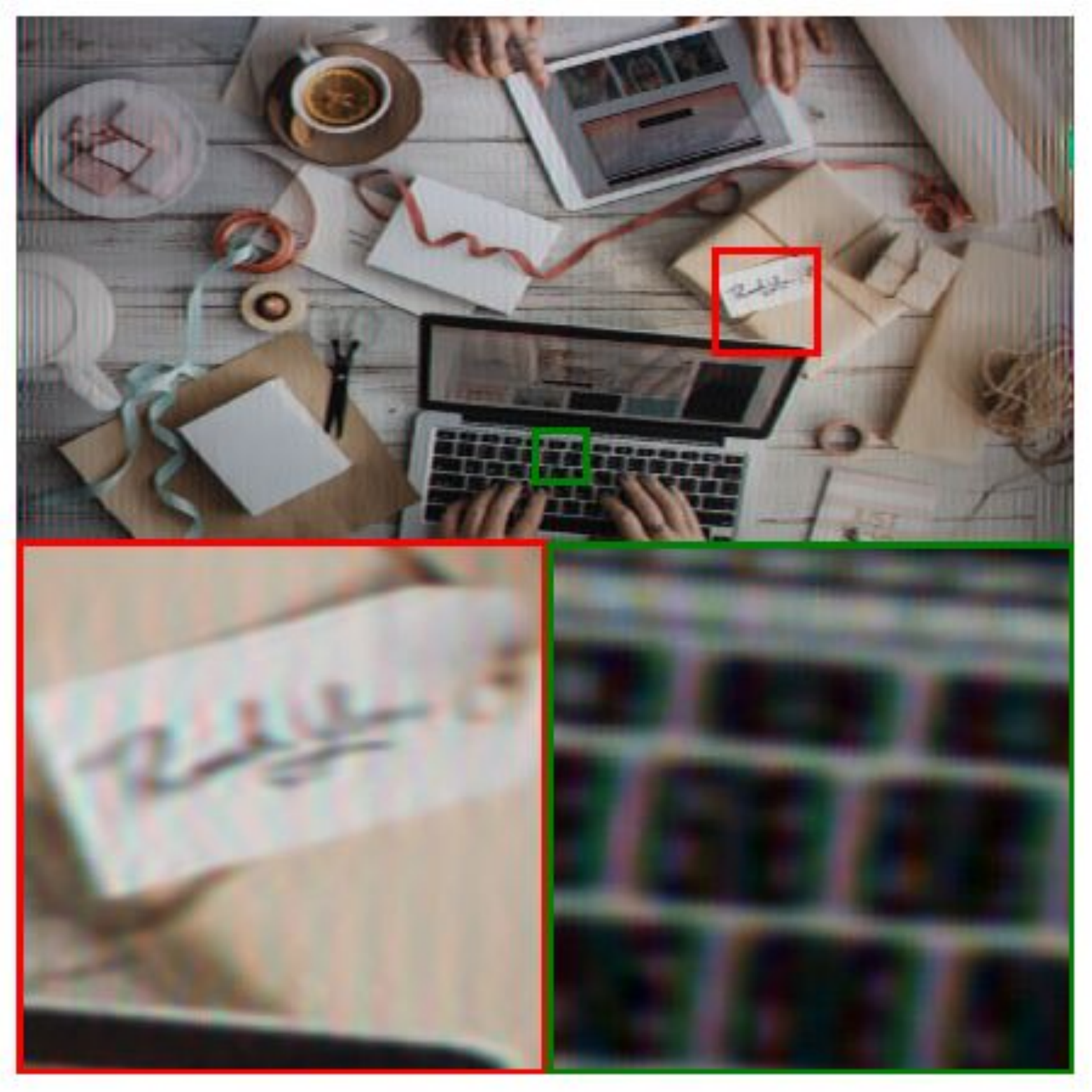}
		\caption{DeP} 
	\end{subfigure}
		\begin{subfigure}{0.19\linewidth} 
		\includegraphics[width=\textwidth]{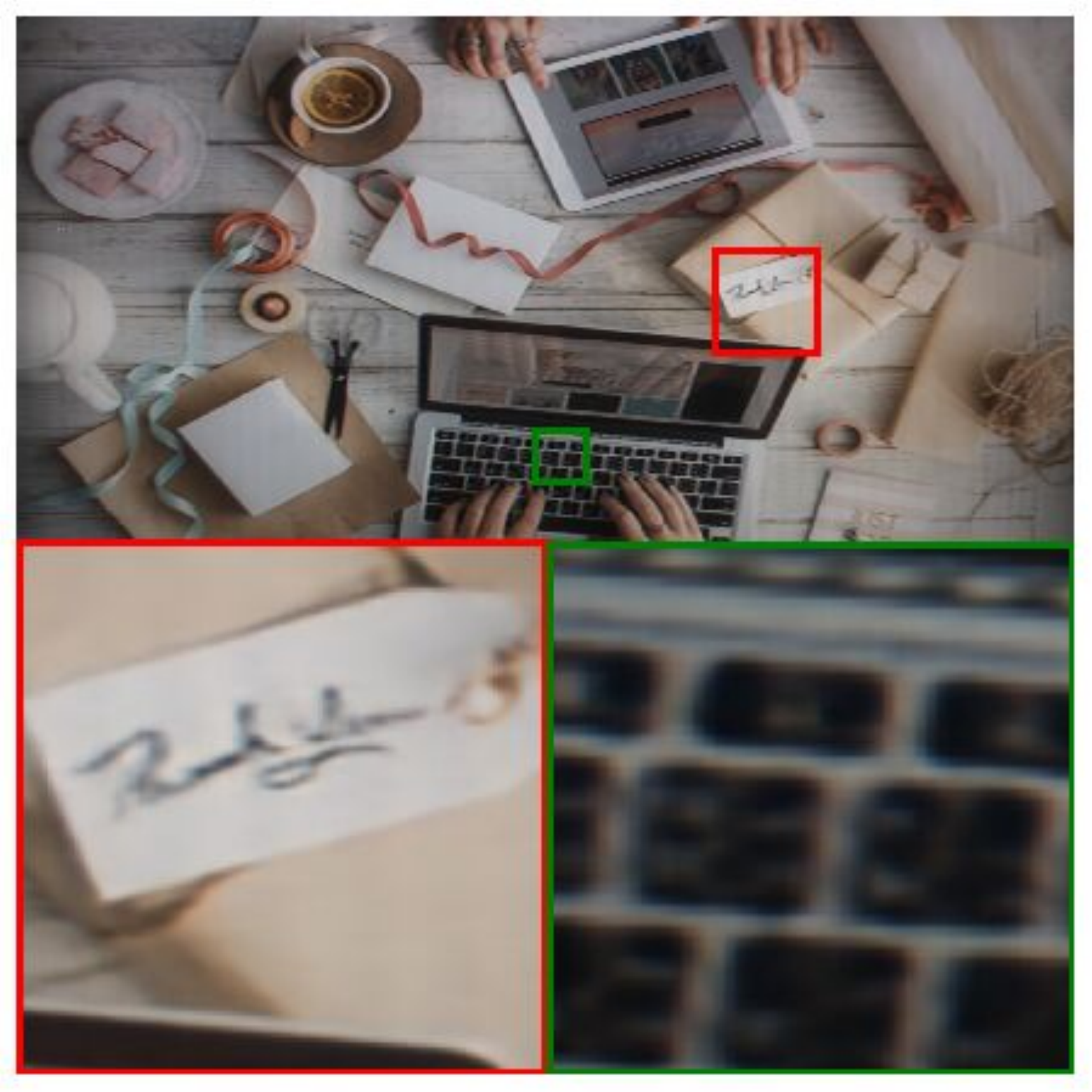}
		\caption{UNet(Syn)} 
	\end{subfigure}
	\begin{subfigure}{0.19\linewidth} 
		\includegraphics[width=\textwidth]{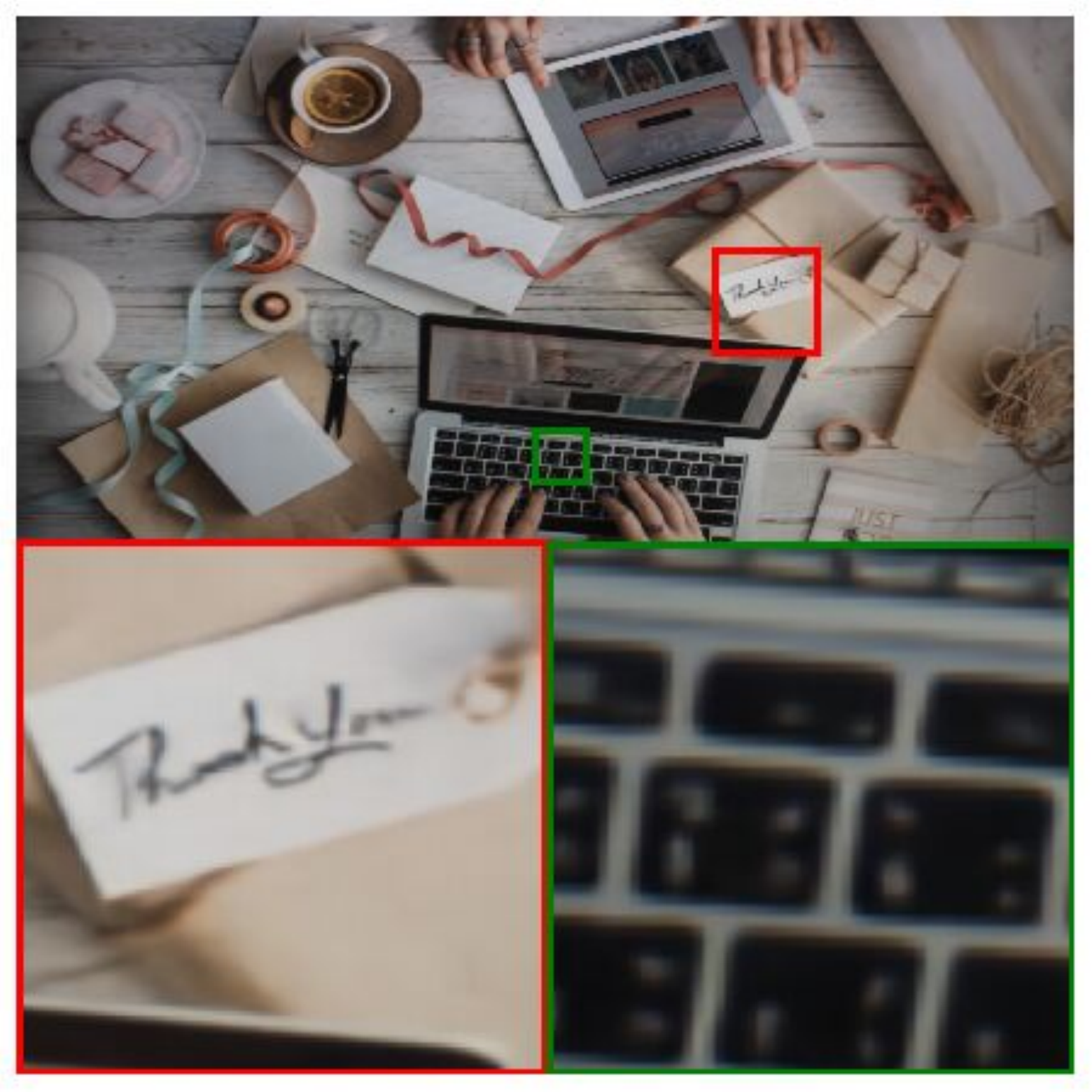}
		\caption{UNet} 
	\end{subfigure}
	\begin{subfigure}{0.19\linewidth} 
		\includegraphics[width=\textwidth]{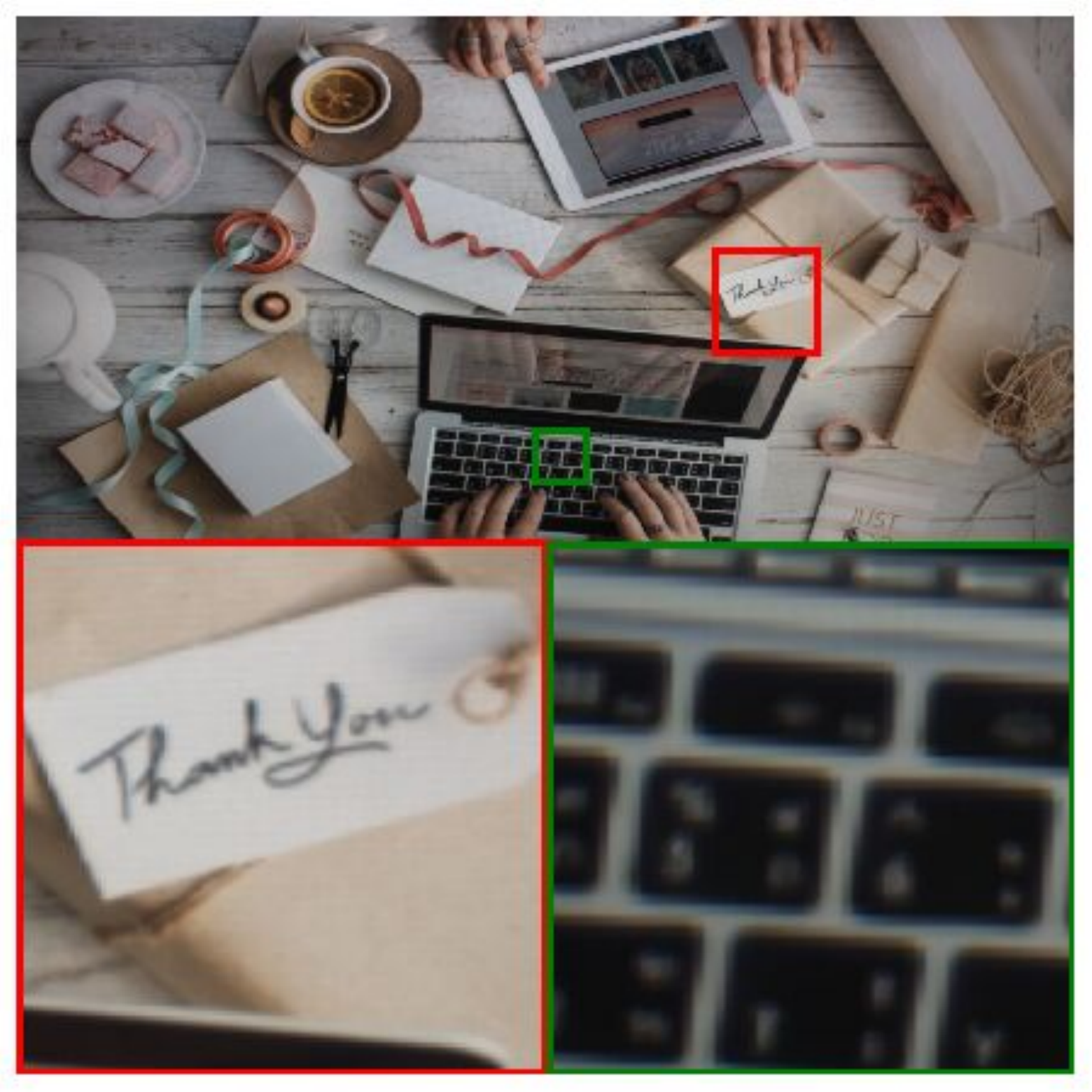}
		\caption{GT} 
	\end{subfigure}
	\caption{Restoration Results Comparison for T-OLED. GT: Ground Truth.}
	\label{fig:toled}
	\vspace{-2mm}
\end{figure*}
\begin{figure*}[t]\setlength{\belowcaptionskip}{-3pt}
\setlength{\abovecaptionskip}{3pt}
	\centering
	\begin{subfigure}{0.19\linewidth} 
		\includegraphics[width=\textwidth]{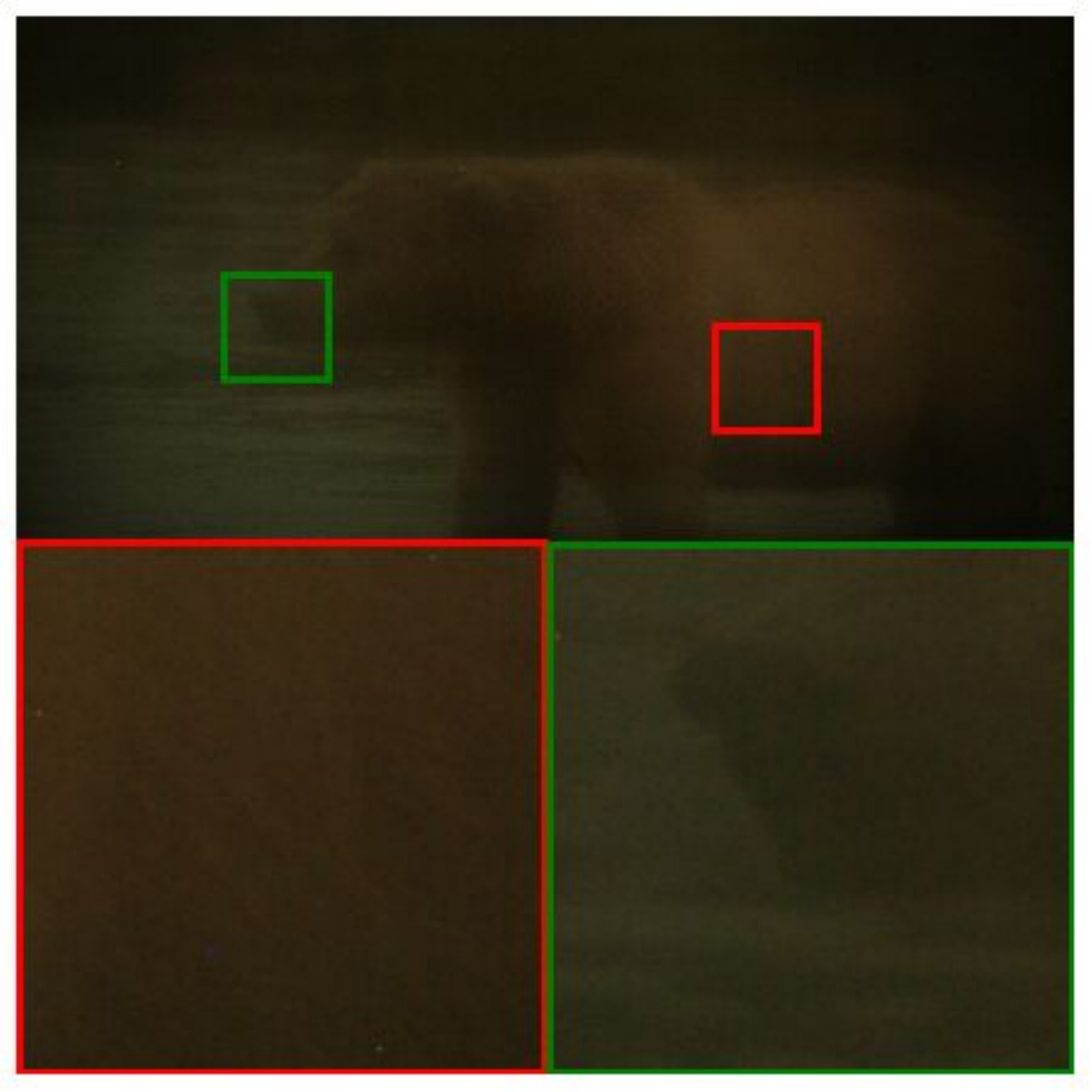}
	\end{subfigure}
	\begin{subfigure}{0.19\linewidth} 
		\includegraphics[width=\textwidth]{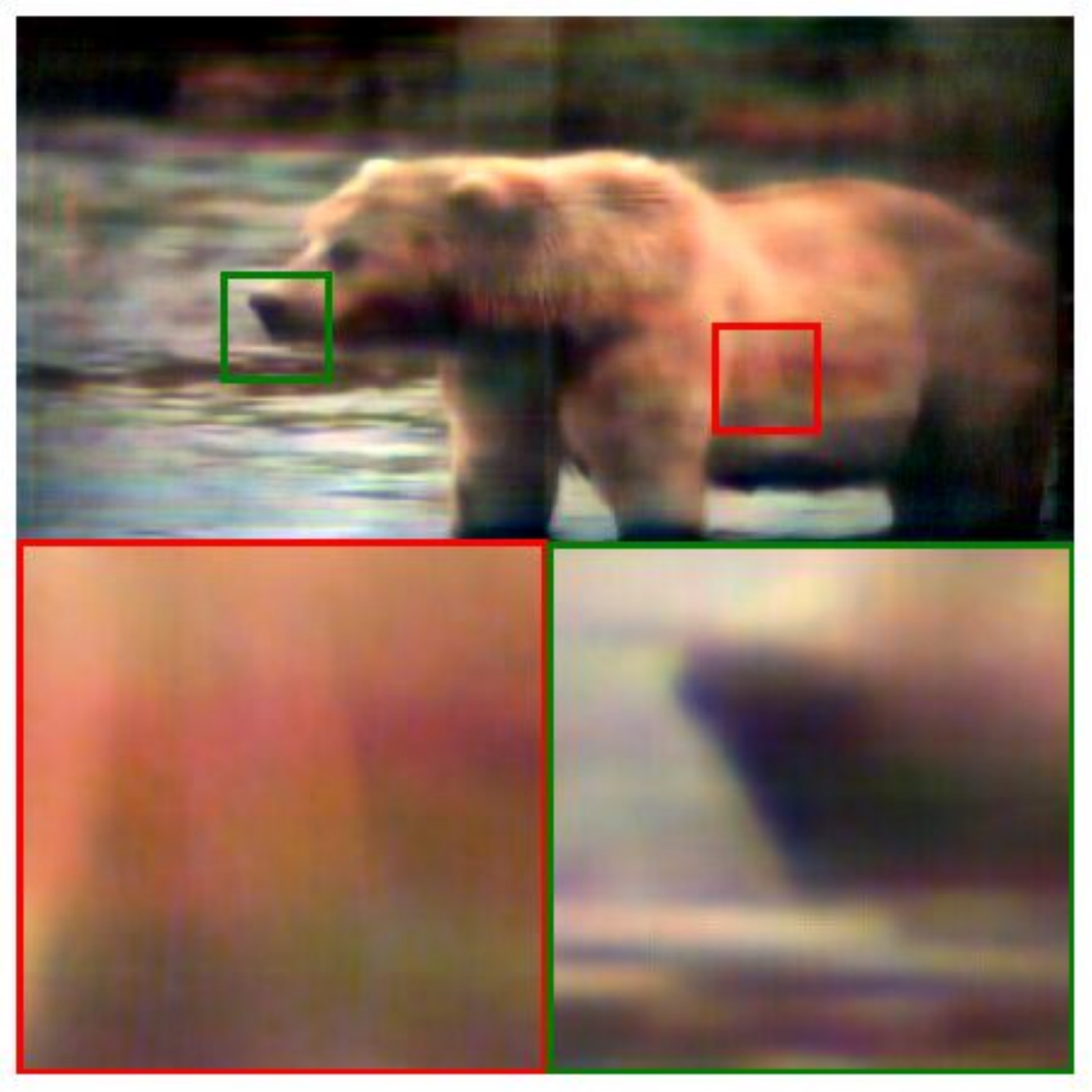}
	\end{subfigure}
		\begin{subfigure}{0.19\linewidth} 
		\includegraphics[width=\textwidth]{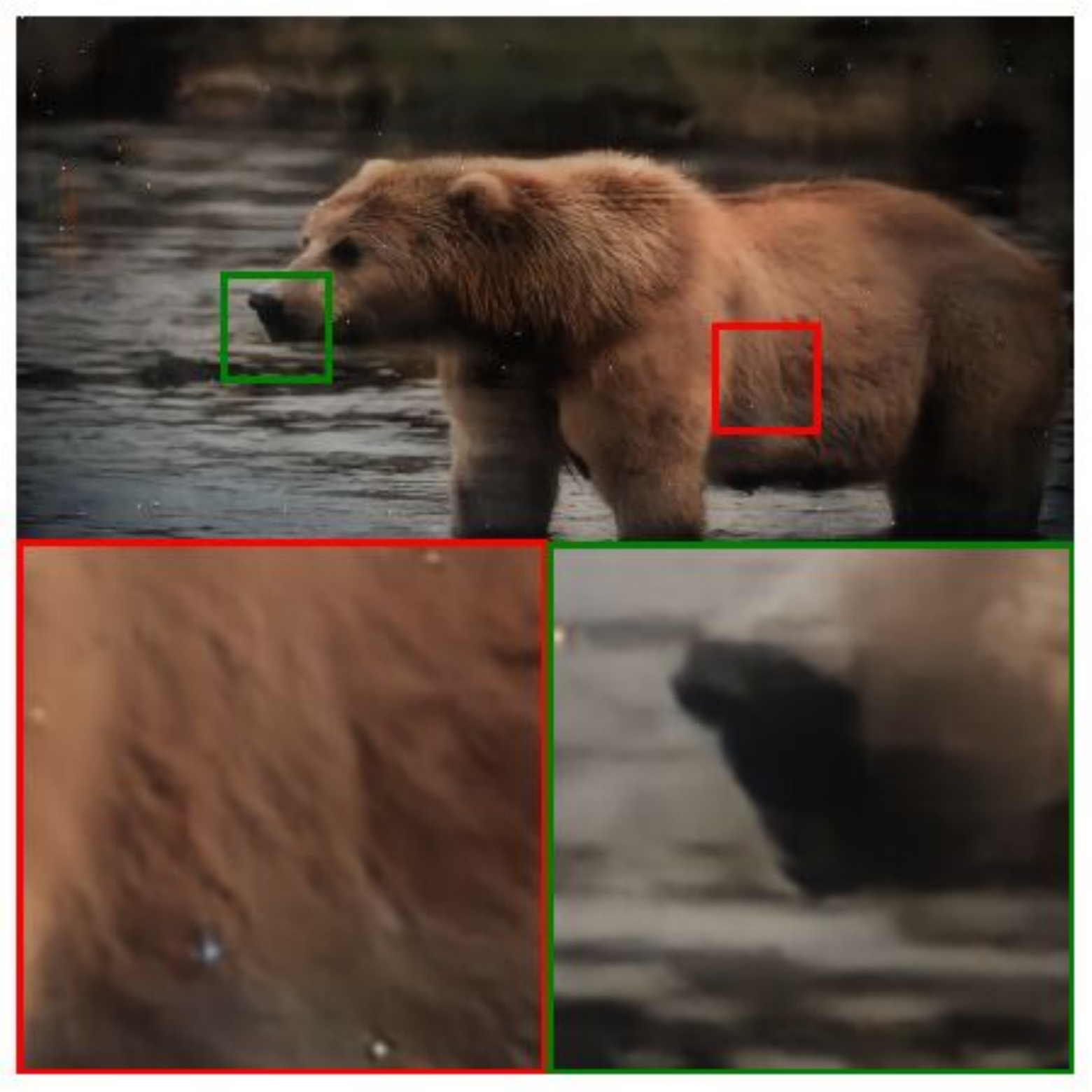}
	\end{subfigure}
	\begin{subfigure}{0.19\linewidth} 
		\includegraphics[width=\textwidth]{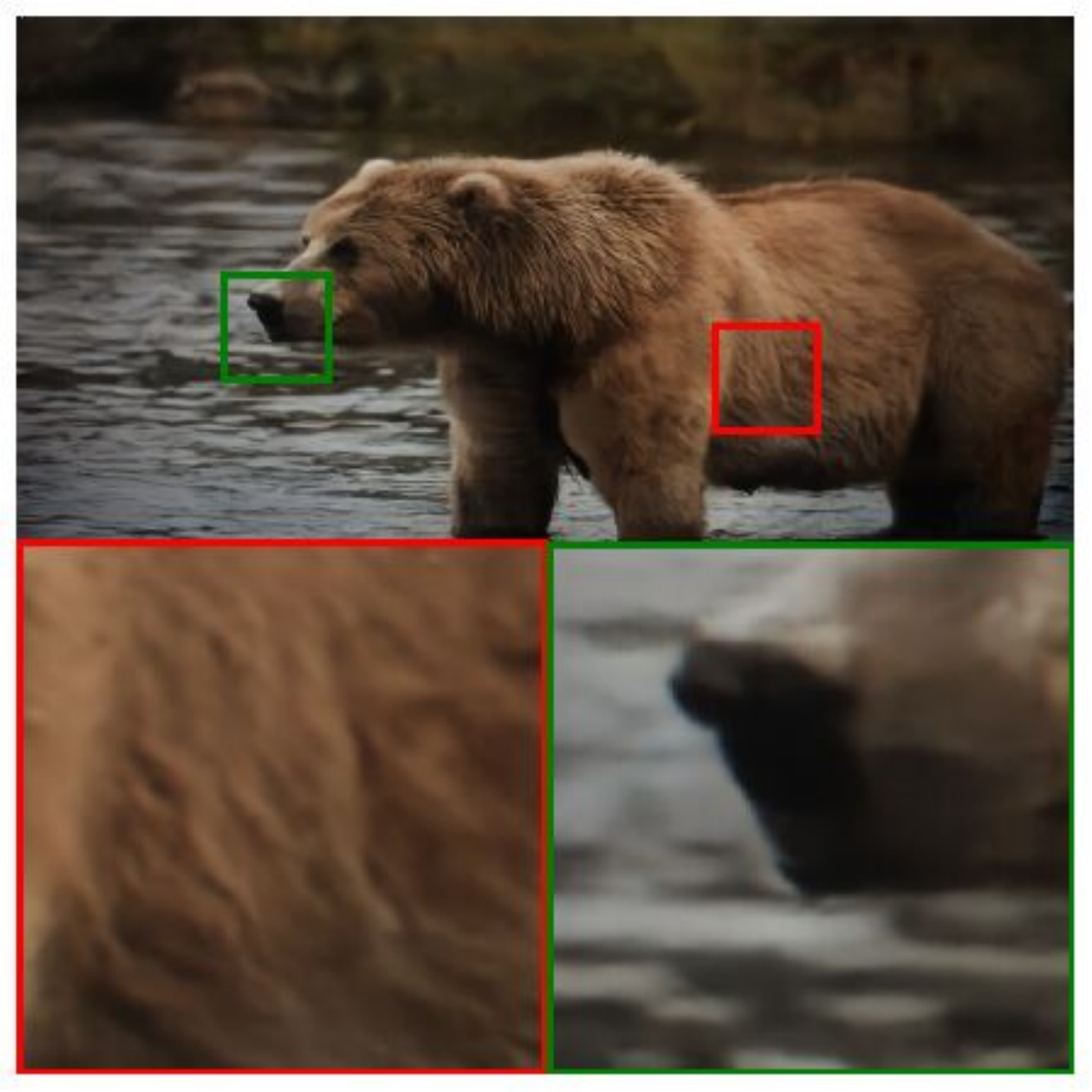}
	\end{subfigure}
	\begin{subfigure}{0.19\linewidth} 
		\includegraphics[width=\textwidth]{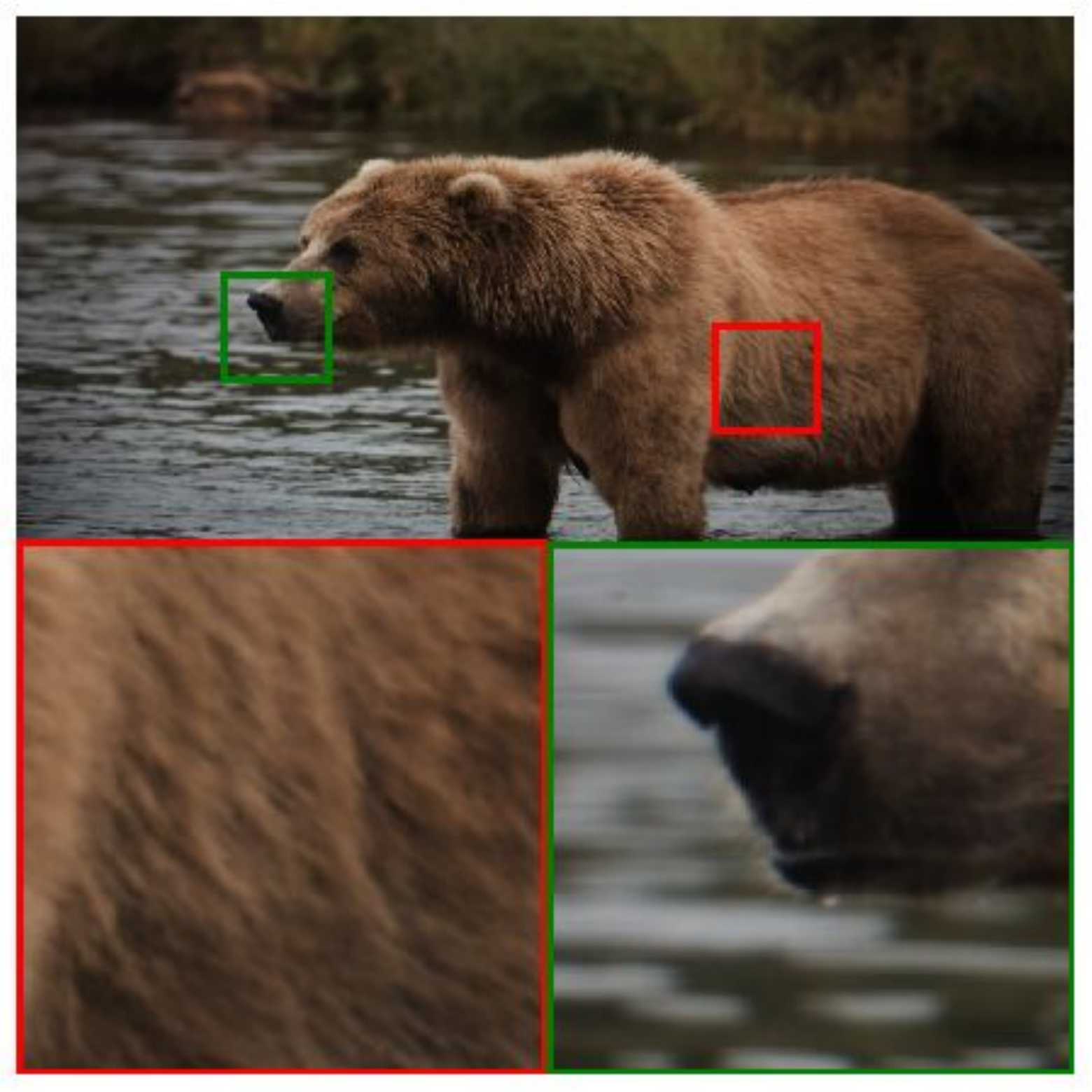}
	\end{subfigure}
	\begin{subfigure}{0.19\linewidth} 
		\includegraphics[width=\textwidth]{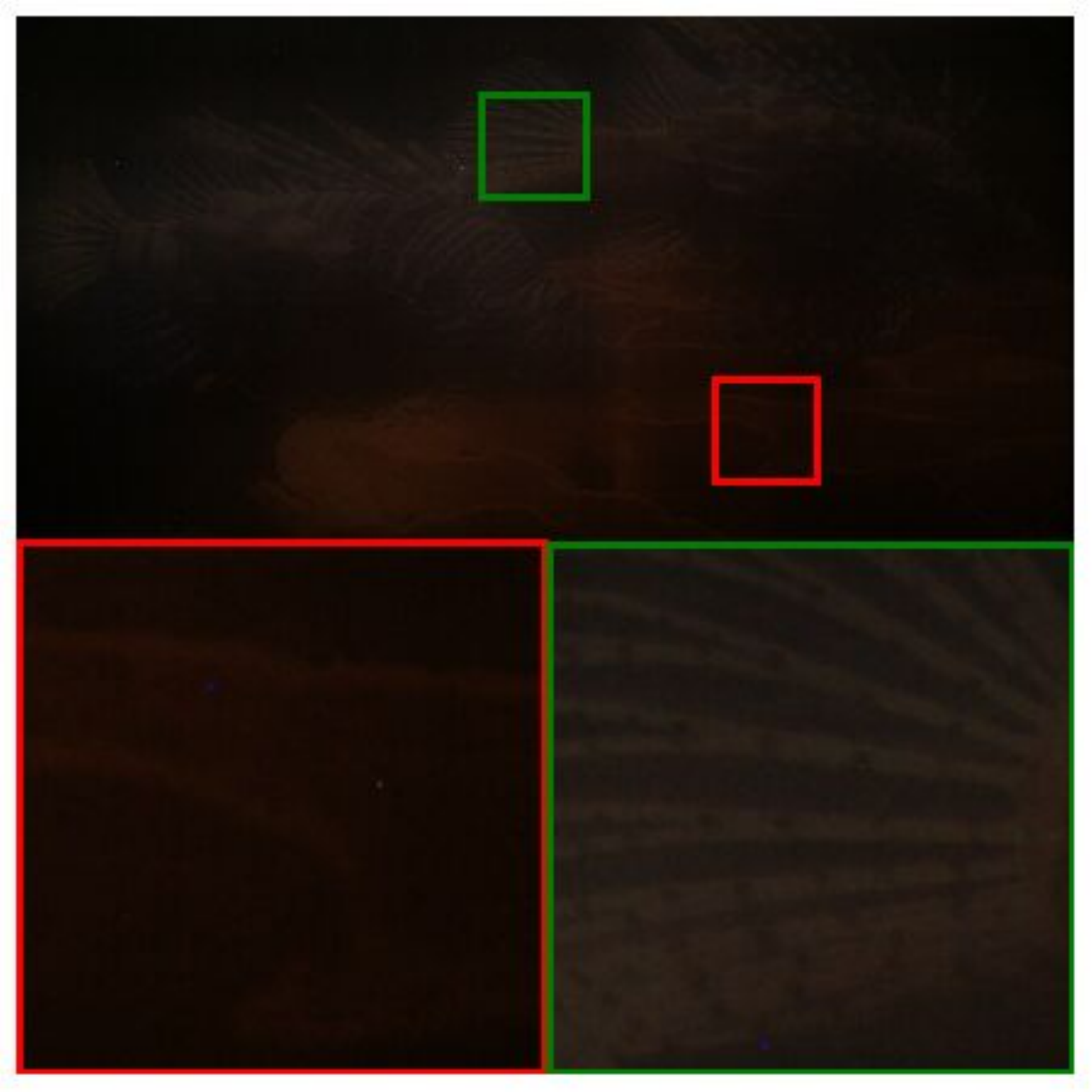}
		\caption{P-OLED} 
	\end{subfigure}
	\begin{subfigure}{0.19\linewidth} 
		\includegraphics[width=\textwidth]{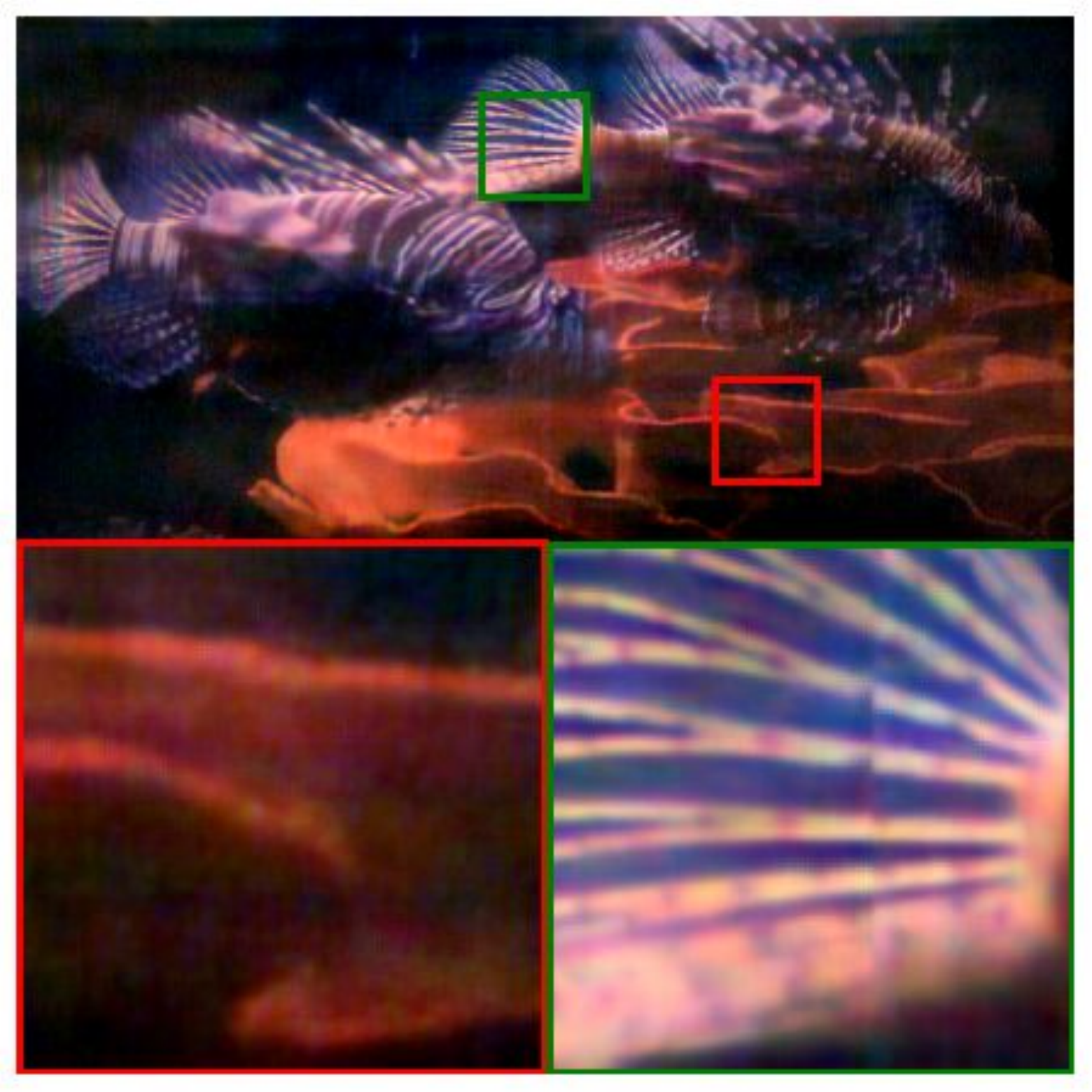}
		\caption{DeP} 
	\end{subfigure}
		\begin{subfigure}{0.19\linewidth} 
		\includegraphics[width=\textwidth]{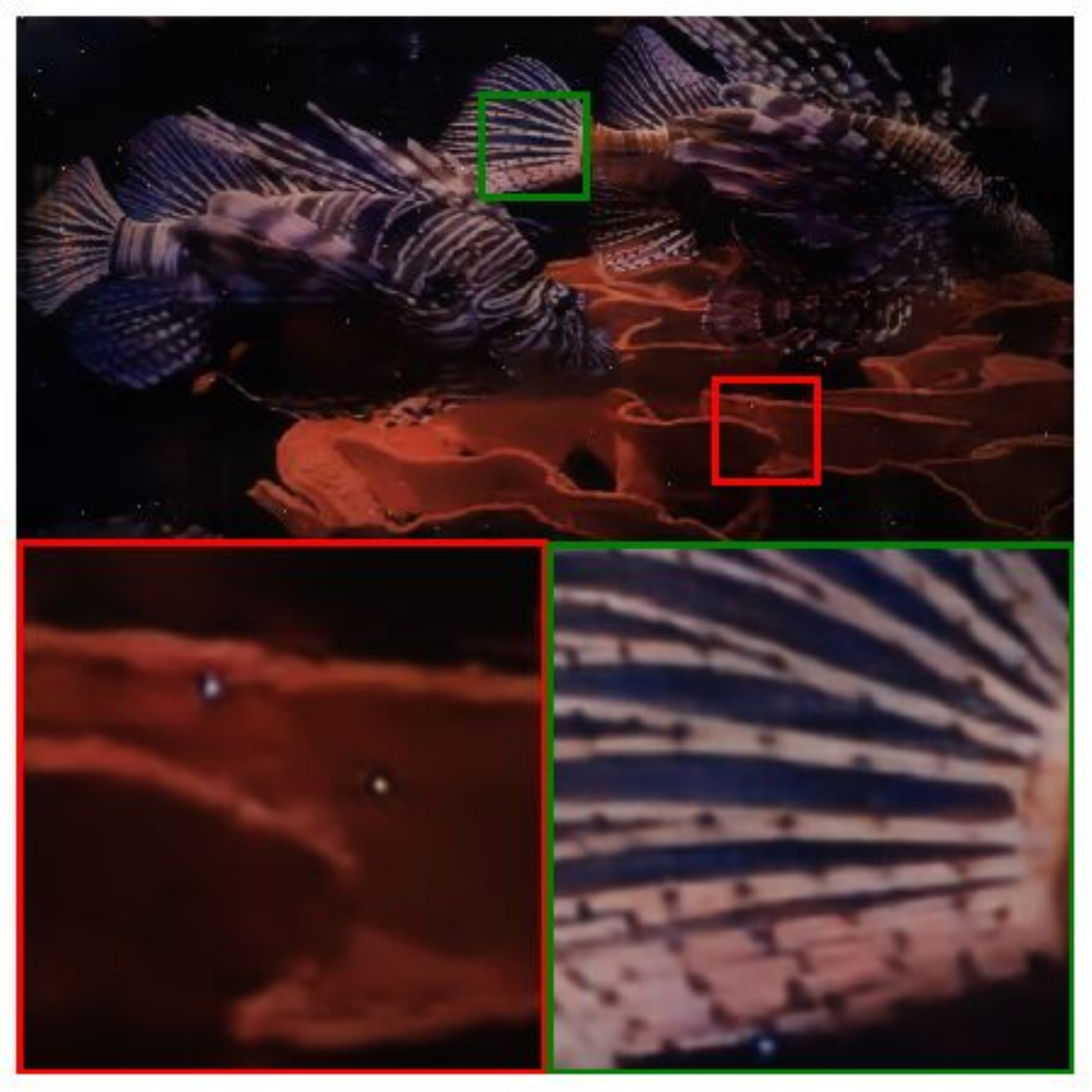}
		\caption{UNet(Syn)} 
	\end{subfigure}
	\begin{subfigure}{0.19\linewidth} 
		\includegraphics[width=\textwidth]{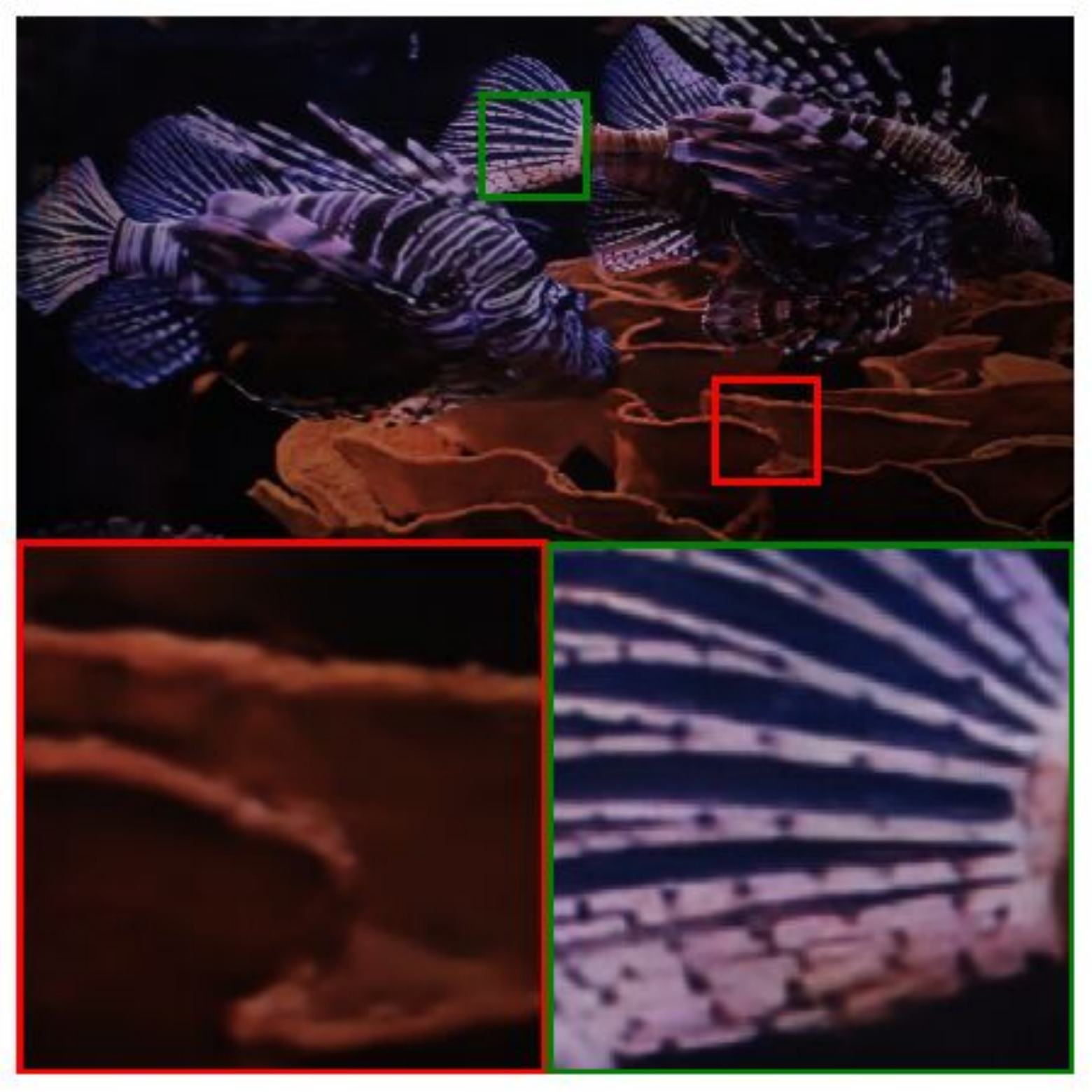}
		\caption{UNet} 
	\end{subfigure}
	\begin{subfigure}{0.19\linewidth} 
		\includegraphics[width=\textwidth]{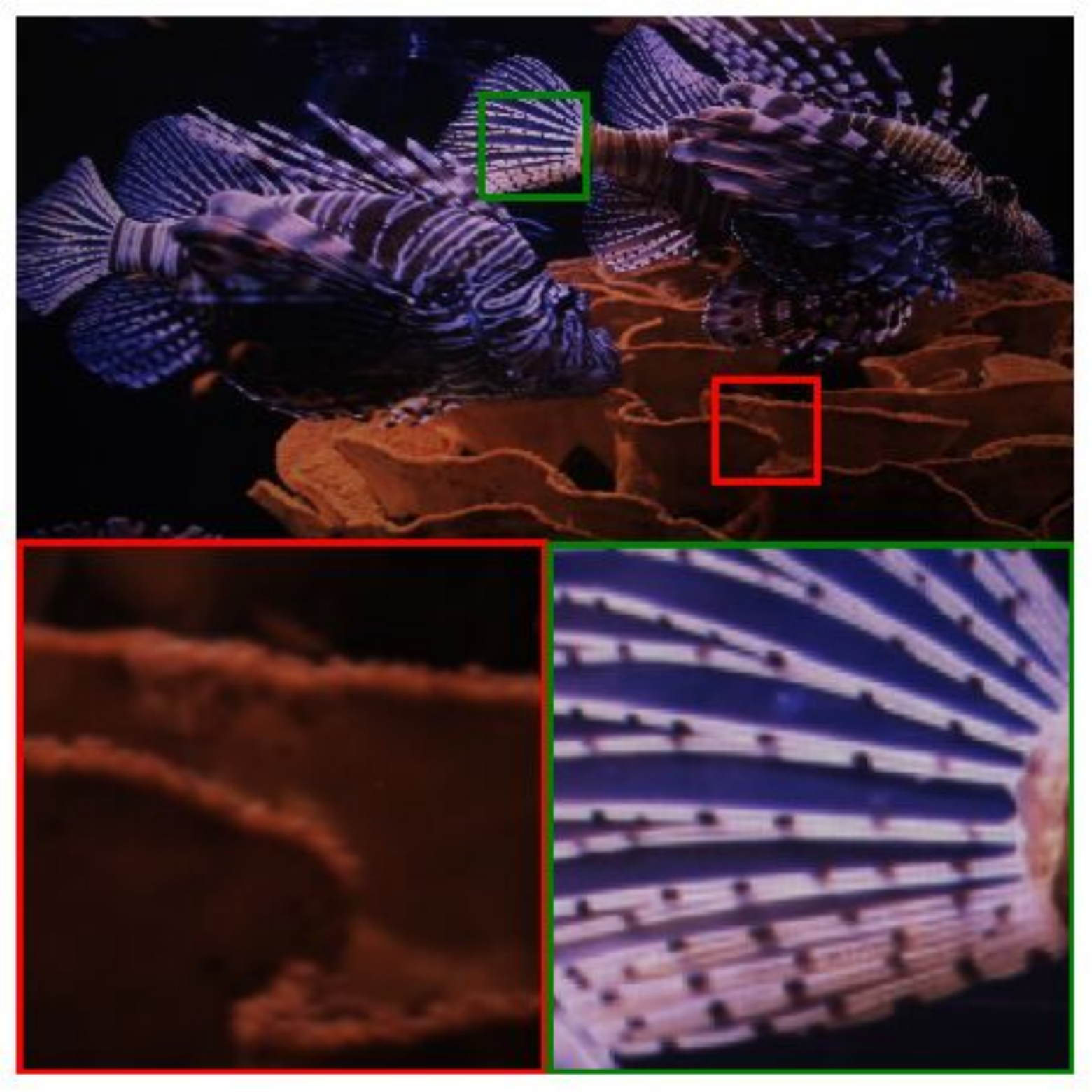}
		\caption{GT} 
	\end{subfigure}
	\caption{Restoration Results Comparison for P-OLED. GT: Ground Truth.}
	\label{fig:poled}
	\vspace{-2mm}
\end{figure*}	
\begin{table*}[t]\setlength{\tabcolsep}{13pt}
\setlength{\abovecaptionskip}{0pt}
\centering
\footnotesize
\caption{Pipeline Comparison}.
\resizebox{\textwidth}{!}{
\begin{tabular}{|l|c|c|c|c|c|c|c|}
\hline
 &&&&\multicolumn{2}{c|}{4K T-OLED} &\multicolumn{2}{c|}{P-OLED} \\\hline\hline
{Pipeline Structure}&{$\#$P} $\downarrow$ &{GFLOPs} $\downarrow$&{T} $\downarrow$ &{PSNR/SSIM} $\uparrow$&{LPIPS} $\downarrow$&{PSNR/SSIM} $\uparrow$&{LPIPS} $\downarrow$\\ \hline
DeP &-&-&-&28.50/0.9117 &0.4219 &16.97/0.7084&0.6306 \\ \hline
ResNet &\bf{1.37M} &721.76 &92.92 &36.26/0.9703 &0.1214 &27.42/0.9176&0.2500\\ \hline
UNet(Syn)  &8.93M&\bf{124.36}&\bf{21.37} &32.42/0.9343 &0.1739&25.88/0.9006 &0.3089 \\ 
UNet &8.93M&\bf{124.36} &\bf{21.37} &\bf{36.71/0.9713}&\bf{0.1209}  &\bf{30.45/0.9427}&\bf{0.2219}\\ \hline
\end{tabular}}
\label{exp:abla2}
\vspace{-4mm}
\end{table*}
\begin{table*}[t]\setlength{\tabcolsep}{12pt}
\setlength{\abovecaptionskip}{0pt}
\centering
\footnotesize
\caption{Ablation Study on UNet alternatives. }
\resizebox{\textwidth}{!}{
\begin{tabular}{|l|c|c|c|c|c|c|c|}
\hline
{Alternatives}&&&&\multicolumn{2}{c|}{4K T-OLED} &\multicolumn{2}{c|}{P-OLED} \\\hline
&{$\#$P} $\downarrow$ &{GFLOPs} $\downarrow$ &{T} $\downarrow$&{PSNR/SSIM} $\uparrow$ &{LPIPS} $\downarrow$ &{PSNR/SSIM} $\uparrow$ &{LPIPS} $\downarrow$ \\ \hline\hline
UNet Basseline&8.93M &124.36 &21.37&36.71/0.9713 &0.1209 &30.45/0.9427 &0.2219\\\hline
Double Width &31.03M&386.37&40.42&37.00/0.9730&0.1171&30.37/0.9425&0.2044\\
Single Encoder&7.76M&97.09&15.85&36.47/0.9704&0.1288&30.26/0.9387&0.2318\\\hline
$L_1 \rightarrow L_1 + SSIM$ &8.93M&124.36&21.37&36.69/0.9714&0.1246&30.37/0.9403&0.2131\\
$L_1 \rightarrow L_1 + VGG$ &8.93M&124.36&21.37&36.31/0.9711&0.1130&30.37/0.9403&0.2130\\
\hline
\end{tabular}}
\label{exp:abla}
\vspace{-4mm}
\end{table*}
\section{Experimental Results}
\subsection{Qualitative and Quantitative Comparisons}
The qualitative restoration results are shown in Figure \ref{fig:toled} and \ref{fig:poled}. As shown, image Deconvolution Pipeline (DeP) successfully recovers image details but still introduces some artifacts, and suffers from the inaccuracy of the computed ideal PSF. The UNet-based model achieves better visual quality and denoising performance. The results of UNet trained with the synthetic data are visually better than DeP. 

The quantitative results are listed in Table \ref{exp:abla2}. We report the performance in PSNR, SSIM, a perceptual metric LPIPS \cite{zhang2018unreasonable}, inference time T (ms/MPixels) and GFLOPs. The inference time is tested with one single Titan X, and the GFLOPs is computed by input size of $512 \times 1024 \times 4$. ResNet achieves a comparable performance to UNet, but it requires more computation operations and longer inference time. The proposed UNet-based structure is efficient and effective, which can therefore be deployed for real-time inference for high-resolution inputs with a single GPU. In Table \ref{exp:abla2}, we demonstrate that synthetic data still has gaps with the real data, though it has already greatly out-performed the DeP for the two display types. The domain gap mainly comes from the following aspects. First, due to the existing distances between display and lens, in real data there appears visible patterns of the display on the image plane. We recall in the assumption of the diffraction model, the display panel is exactly at the principle plane of the lens system. The cause of the visible bands are illustrated in the supplementary material. Second, the approximated light transmission rate may not be accurate, the measured values may be influenced by other environment light sources. Third, impulse noise caused by dead pixels or over-exposure in the camera sensors widely exist in the real dataset. Those factors provide more improvement space for this work.
\begin{figure}[t]
	\centering
		\includegraphics[width=\linewidth]{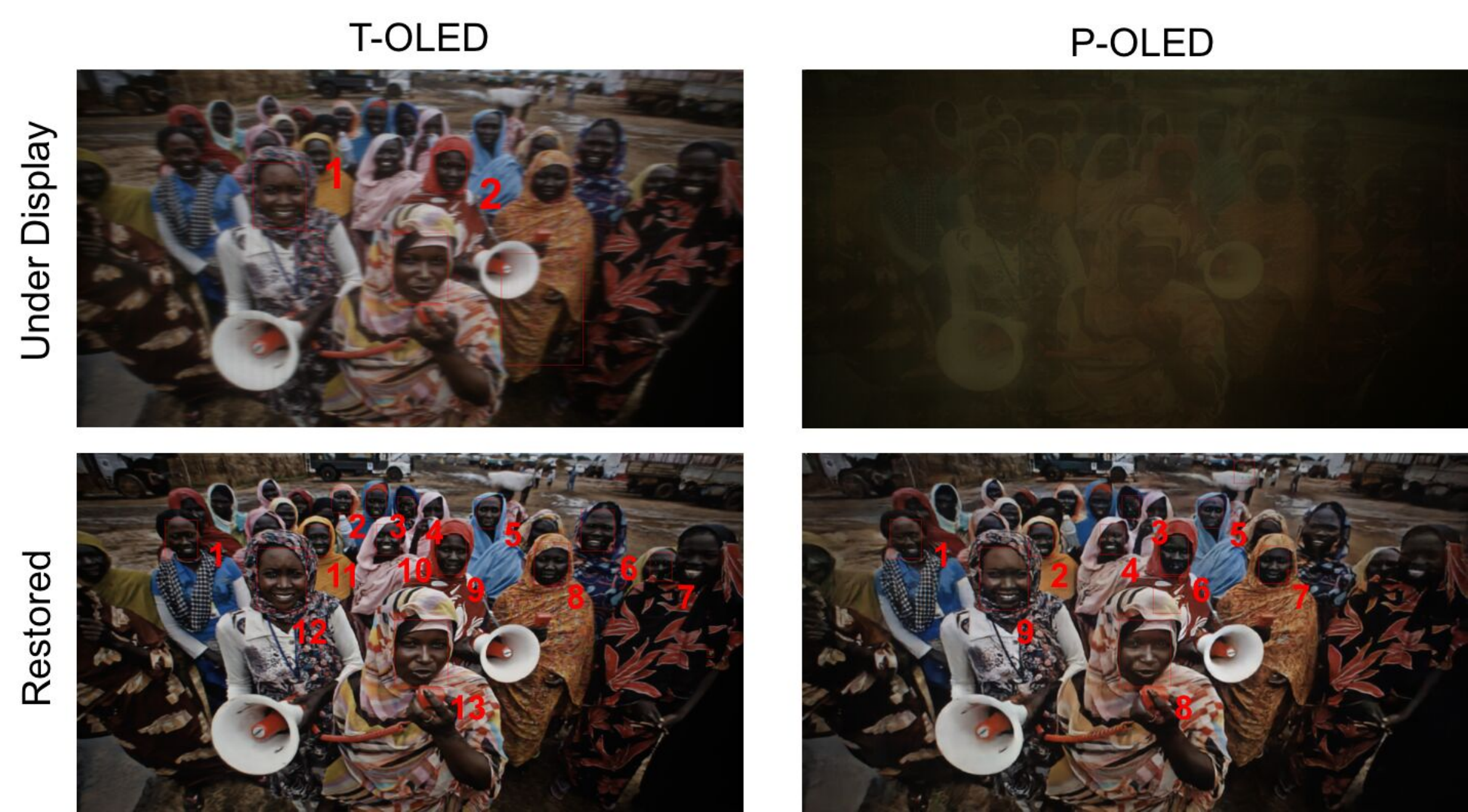}
	\caption{Face detection performance before and after applying restoration. Without display, the original face recall rate is 60$\%$. Covering the camera with T-OLED or P-OLED will decrease the recall rate to 8$\%$ and 0$\%$. After image restoration, the recall rates recovered back to 56$\%$ and 39$\%$.}
	\label{fig:fd}
	\vspace{-4mm}
\end{figure}
\subsection{Ablation Study}
For the best-performed UNet structure, we compare different UNet alternatives in Table \ref{exp:abla}. We increase the parameter size by splitting the original encoders into two sub-encoders, so the performance is also increased. The increment parameter size and inference time is far less than doubling the width of each layer of UNet, but the performance improvement is comparable (T-OLED), even better (P-OLED). We claim that the proposed UNet structure will both maintain a small number of parameters and operations, and achieve a real-time high-quality inference. To try alternative loss functions, we add $SSIM$ or $VGG$ loss in additional to $L_1$ loss with 1:1 ratio. However, the performance gains on either $SSIM$ or perceptual metric LPIPS are not significant enough, and are not visually distinctive. Adversarial loss is not implemented due to its temporal instability of GAN-based training. 

\subsection{Downstream Applications}
The proposed image restoration also enhances the performance of downstream applications including face detection. Figure \ref{fig:fd} shows an example of detecting faces using MTCNN \cite{zhang2016joint}. Without display, the original face recall rate is 60$\%$. Covering the camera with T-OLED or P-OLED will decrease the recall rate to 8$\%$ and 0$\%$. After image restoration, the recall rates are recovered to 56$\%$ and 39$\%$.

\section{Conclusion and Limitations}
This paper defined and presented a novel imaging system named Under-Display-Camera (UDC). Deploying UDC to full-screen devices improves the user interaction as well as teleconferencing experience, but does harm to imaging quality and other downstream vision applications. We systematically analyzed the optical systems and modelled the image formation pipeline of UDC, and both collected real data using a novel acquisition system and synthesized realistic data and the PSF of the system using optical model. We then proposed to address the image restoration of UDC using a Deconvolution-based Pipeline (DeP) and data-driven learning-based methods. Our experiments showed that the former achieved basic restoration and the latter demonstrated an efficient high-quality restoration. The model trained with synthetic data also achieved a remarkable performance indicating the potential generalization ability. 

UDC problem has its promising research values in complicated degradation analysis. In real-world applications, other factors like an active display, reflection, lens flare \textit{etc.} are still very challenging and complicated. Future work can be exploring UDC-specific restoration models and working with aperture and display researchers to analyze the influential factors of image degradation. It will make the restoration model generalized better for mass production, or helpful for down-stream tasks, as an ultimate goal.

\appendix
\section{Appendices}
\renewcommand\thefigure{\thesection.\arabic{figure}}    
\setcounter{figure}{0} 

\begin{figure*}[t]
	\centering
		\begin{subfigure}{0.33\linewidth} 
		\includegraphics[width=\textwidth]{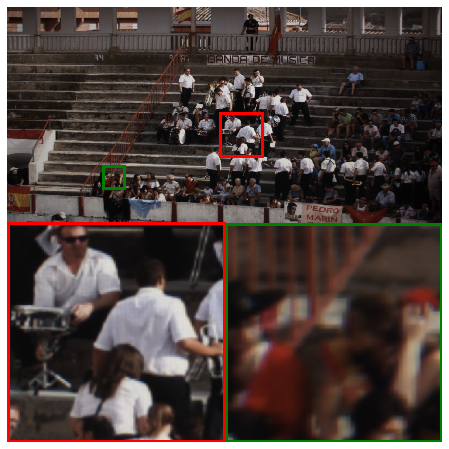}
		\caption{Display-free}
	\end{subfigure}
	\begin{subfigure}{0.33\linewidth} 
		\includegraphics[width=\textwidth]{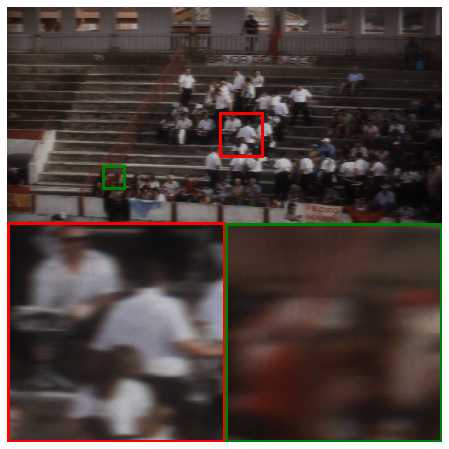}
		\caption{T-OLED}
	\end{subfigure}
		\begin{subfigure}{0.33\linewidth} 
		\includegraphics[width=\textwidth]{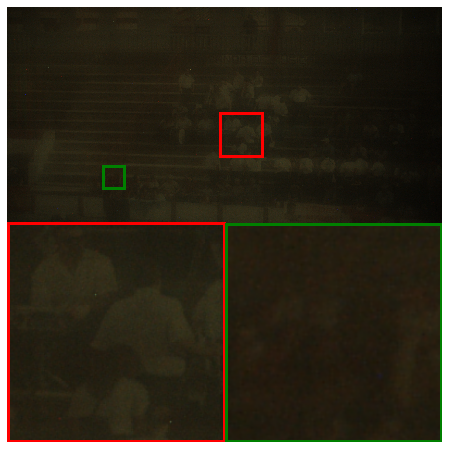}
		\caption{P-OLED}
	\end{subfigure}
	\caption{More real data samples acquried by our MCIS set-up. (a) The image captured with camera covered by thin glass, (b)T-OLED, and (c) P-OLED.}
	\label{fig:data}
\end{figure*}

\subsection{Real Data}

More examples in 8-bit RGB version in the UDC real dataset are shown in Fig. \ref{fig:data}. Each image has a high resolution of $1024 \times 2048 \times 3$. Images captured by T-OLED demonstrate a blur effect along the horizontal direction. Some spatial frequencies (i.e. vertical bands) are missing due to diffraction effects. Images captured by P-OLED are yellow-shifted, dark, and noisy. We also stored the 16-bit raw sensor data, which is mainly used for training and testing in the paper. 

\subsection{Synthetic Data}
\begin{figure}[t]
	\centering
		\includegraphics[width=\columnwidth]{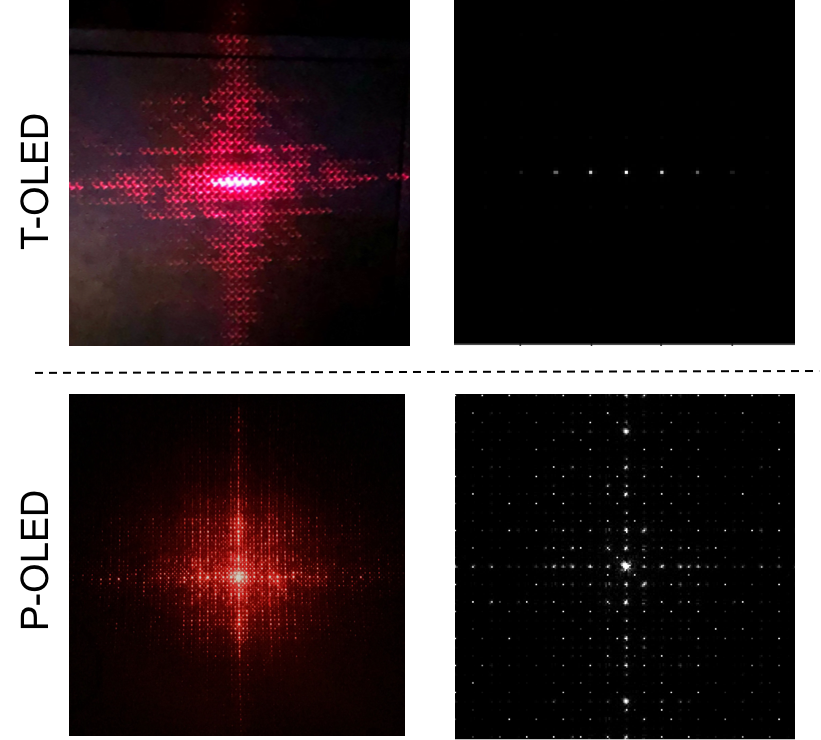}
	\caption{Real and the computed point spread function (kernel).}
	\label{fig:PSF}
\end{figure}

\begin{figure}[h]
	\centering
		\begin{subfigure}{0.49\linewidth} 
		\includegraphics[width=\textwidth]{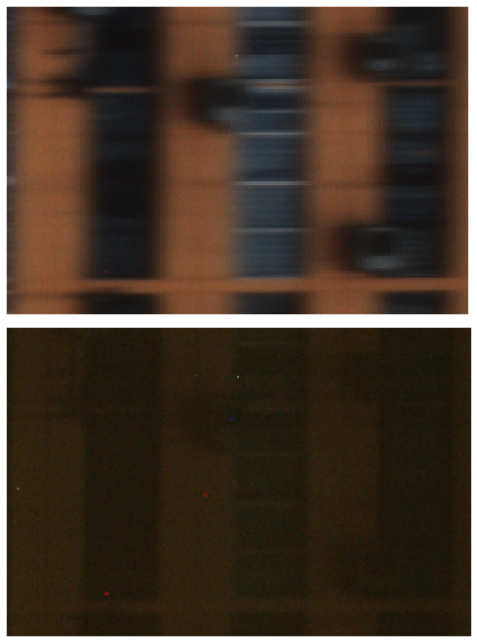}
		\caption{Real data samples.}
	\end{subfigure}
	\begin{subfigure}{0.49\linewidth} 
		\includegraphics[width=\textwidth]{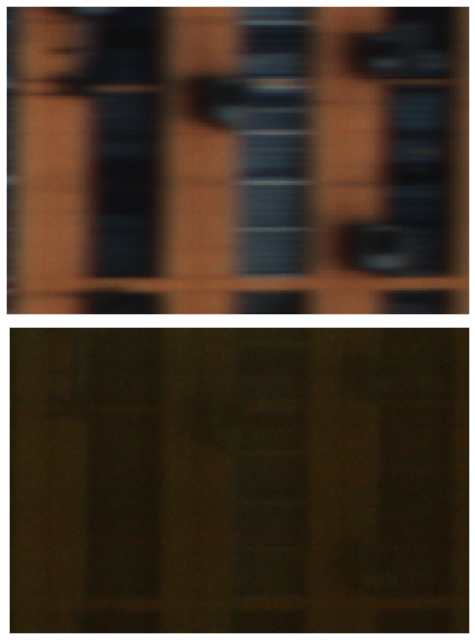}
		\caption{Synthetic data samples.}
	\end{subfigure}
	\caption{Comparison of real data and synthetic data. First row: T-OLED. Second row: P-OLED.}
	\label{fig:comprs}
\end{figure}

We follow the image formation pipeline to synthesize the near-realistic data. Given only the display pattern, and some specific measurements of the cameras, we could generate the blur kernels as shown in Fig. \ref{fig:PSF} along with the degraded images for training. Fig. \ref{fig:comprs} compares the synthetic data with the real data. Perceptually, two sets of data samples have similar visual characteristics.  

\subsection{Visible Bands for T-OLED}

\begin{figure}[t]
	\centering
		\begin{subfigure}{0.49\linewidth} 
		\includegraphics[width=\textwidth]{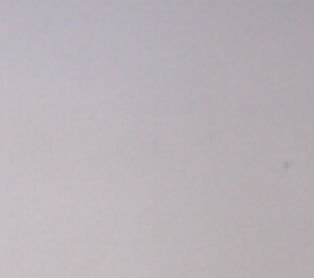}
		\caption{Synthetic data.}
	\end{subfigure}
	\begin{subfigure}{0.49\linewidth} 
		\includegraphics[width=\textwidth]{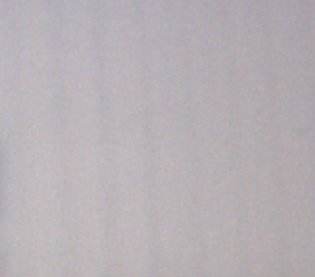}
		\caption{Real data with bands.}
	\end{subfigure}
	\caption{Visible bands in real data.}
	\label{fig:syntoled}
\end{figure}

In addition to the degradation formulated in the paper, there is another minor image artifact caused by the periodic grating-like pixel structure (i.e. T-OLED): superposition of periodic bands over the image at low to moderate visibility levels. As shown in the Fig. \ref{fig:syntoled}, periodic bands are visible in the real data, but not in the synthetic data. We regard it as the main gap of data synthesis. 
Those bands are caused by the imperfect adhesion of the display to the camera lens. In the degradation model, we assume the display pattern or objects are exactly placed against the lens, while in practical set-up of our experiments, there is still a small distance between them. We can consider the grating as being imaged very out-of-focus on the sensor plane. There will be an image on the image sensors consisting of the grating convoluted with the very-out-of-focus point spread function – a circle.
This problem can be mitigated by real industrial manufacturing process, so we did not resolve it explicitly in the paper with experimental settings. However, it still forms an interesting problem regarding eliminating real periodic noises left for future works.

\subsection{More Restoration Results}
\begin{figure*}[t]
	\centering
	\begin{subfigure}{0.19\linewidth} 
		\includegraphics[width=\textwidth]{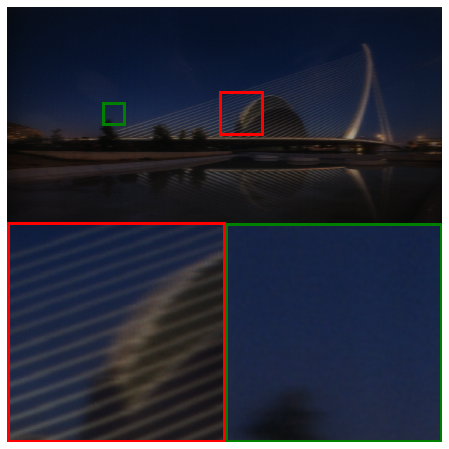}
	\end{subfigure}
	\begin{subfigure}{0.19\linewidth} 
		\includegraphics[width=\textwidth]{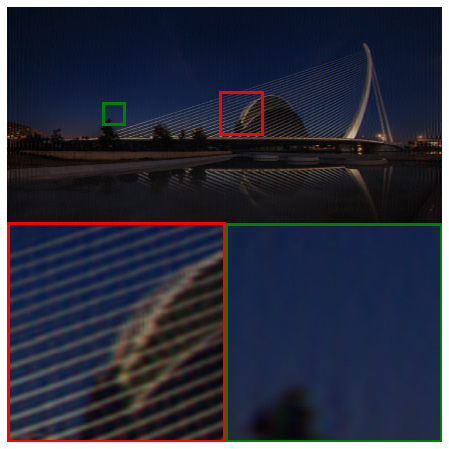}
	\end{subfigure}
		\begin{subfigure}{0.19\linewidth} 
		\includegraphics[width=\textwidth]{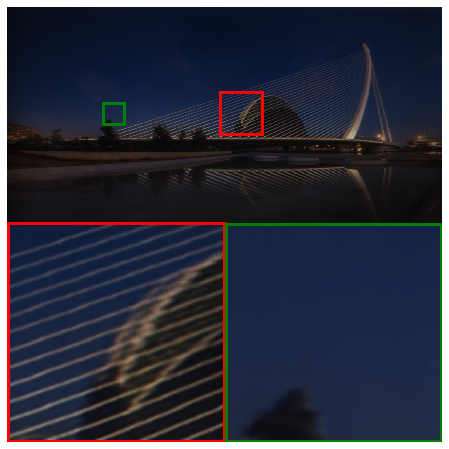}
	\end{subfigure}
	\begin{subfigure}{0.19\linewidth} 
		\includegraphics[width=\textwidth]{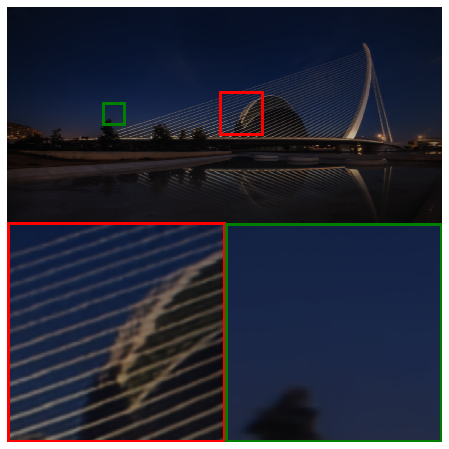}
	\end{subfigure}
	\begin{subfigure}{0.19\linewidth} 
		\includegraphics[width=\textwidth]{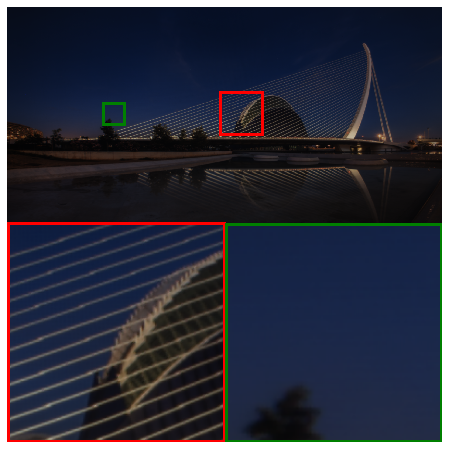}
	\end{subfigure}
	
	\begin{subfigure}{0.19\linewidth} 
		\includegraphics[width=\textwidth]{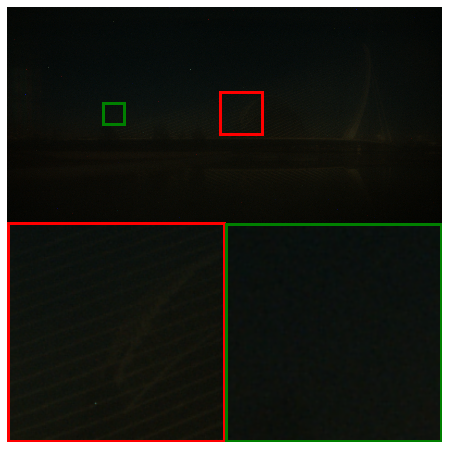}
	\end{subfigure}
		\begin{subfigure}{0.19\linewidth} 
		\includegraphics[width=\textwidth]{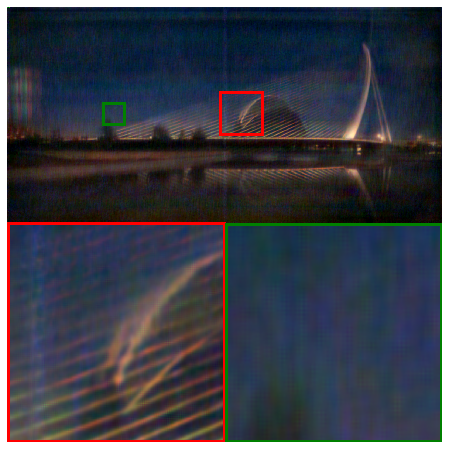}
	\end{subfigure}
		\begin{subfigure}{0.19\linewidth} 
		\includegraphics[width=\textwidth]{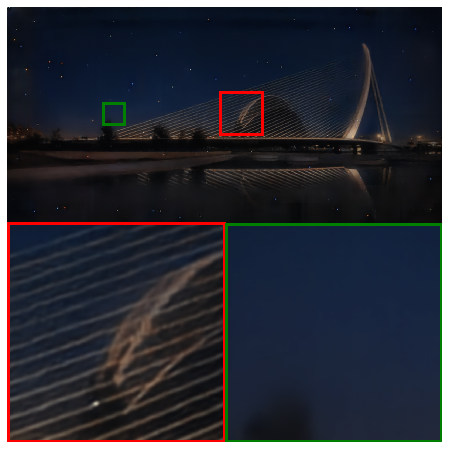}
	\end{subfigure}
	\begin{subfigure}{0.19\linewidth} 
		\includegraphics[width=\textwidth]{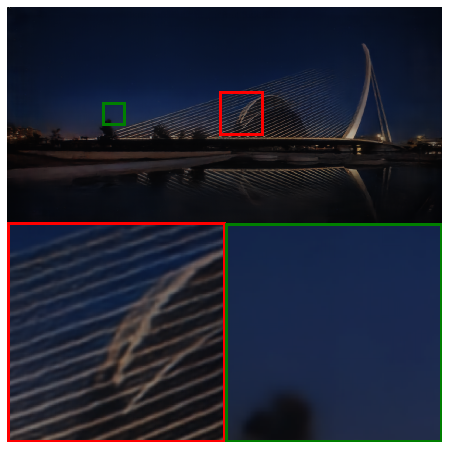}
	\end{subfigure}
	\begin{subfigure}{0.19\linewidth} 
		\includegraphics[width=\textwidth]{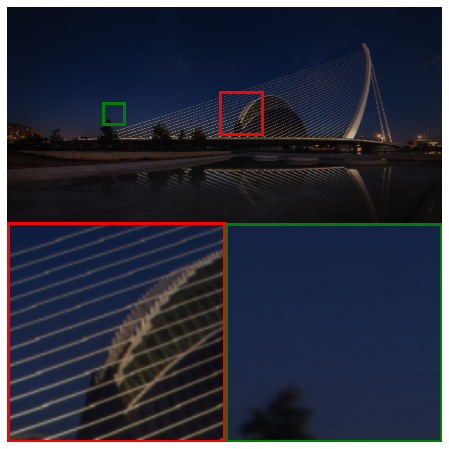}
	\end{subfigure}
		
	\begin{subfigure}{0.19\linewidth} 
		\includegraphics[width=\textwidth]{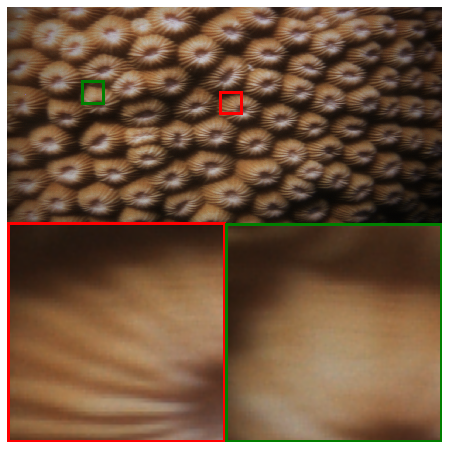}
	\end{subfigure}
	\begin{subfigure}{0.19\linewidth} 
		\includegraphics[width=\textwidth]{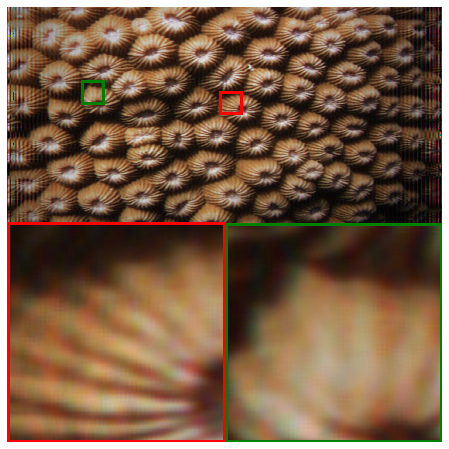}
	\end{subfigure}
		\begin{subfigure}{0.19\linewidth} 
		\includegraphics[width=\textwidth]{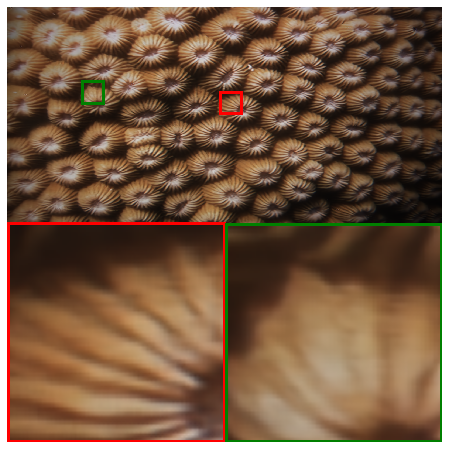}
	\end{subfigure}
	\begin{subfigure}{0.19\linewidth} 
		\includegraphics[width=\textwidth]{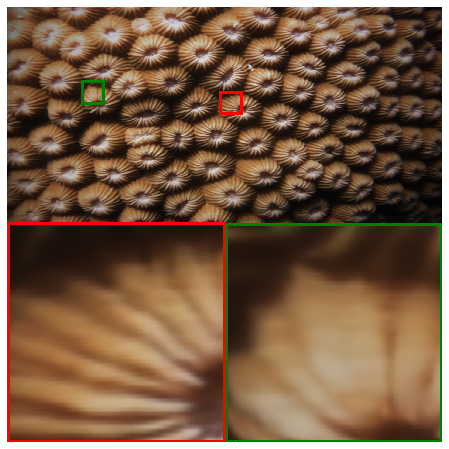}
	\end{subfigure}
	\begin{subfigure}{0.19\linewidth} 
		\includegraphics[width=\textwidth]{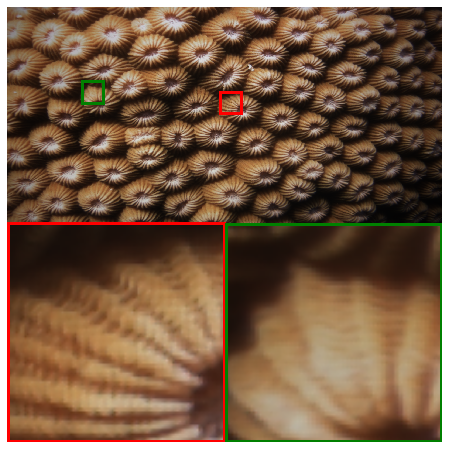}
	\end{subfigure}
	
	\begin{subfigure}{0.19\linewidth} 
		\includegraphics[width=\textwidth]{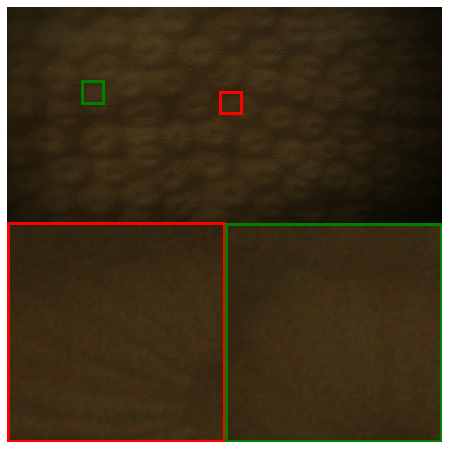}
	\end{subfigure}
	\begin{subfigure}{0.19\linewidth} 
		\includegraphics[width=\textwidth]{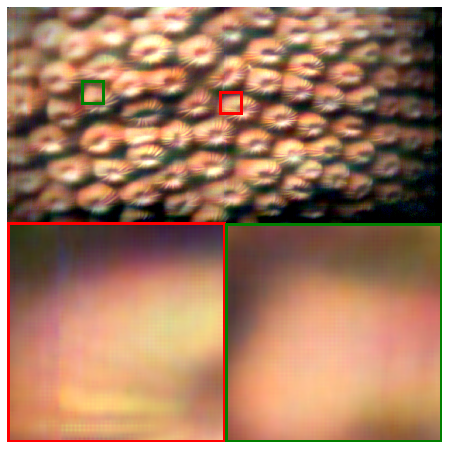}
	\end{subfigure}
		\begin{subfigure}{0.19\linewidth} 
		\includegraphics[width=\textwidth]{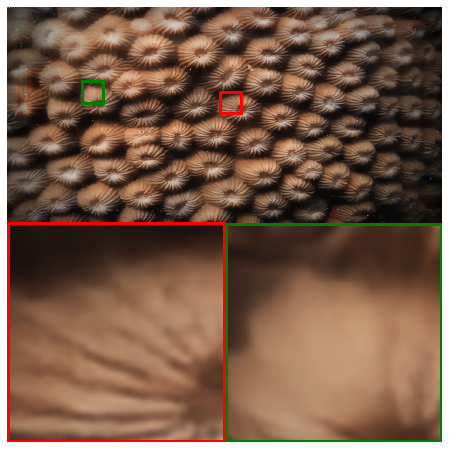}
	\end{subfigure}
	\begin{subfigure}{0.19\linewidth} 
		\includegraphics[width=\textwidth]{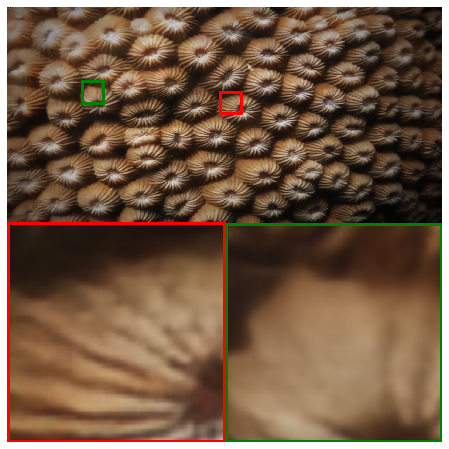}
	\end{subfigure}
	\begin{subfigure}{0.19\linewidth} 
		\includegraphics[width=\textwidth]{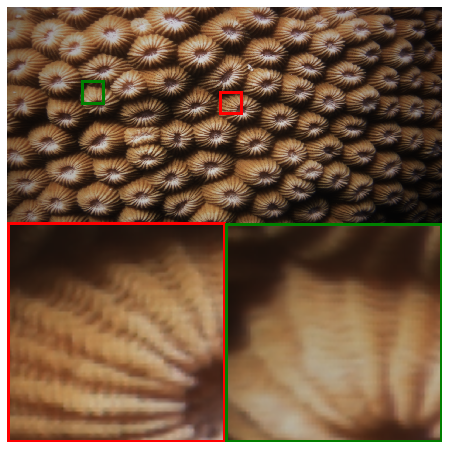}
	\end{subfigure}

	\begin{subfigure}{0.19\linewidth} 
		\includegraphics[width=\textwidth]{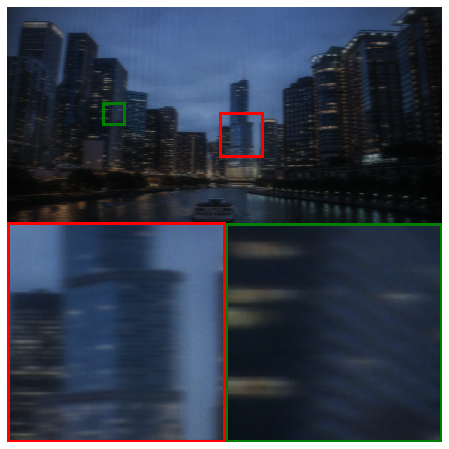}
	\end{subfigure}
	\begin{subfigure}{0.19\linewidth} 
		\includegraphics[width=\textwidth]{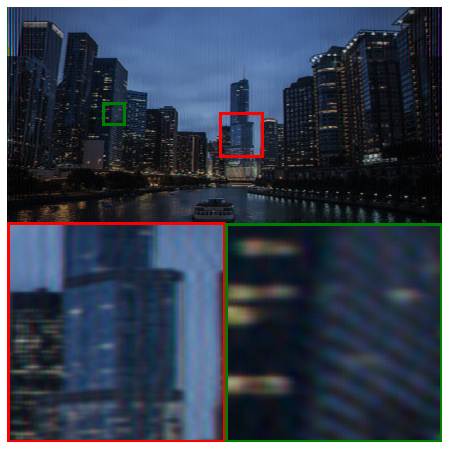}
	\end{subfigure}
		\begin{subfigure}{0.19\linewidth} 
		\includegraphics[width=\textwidth]{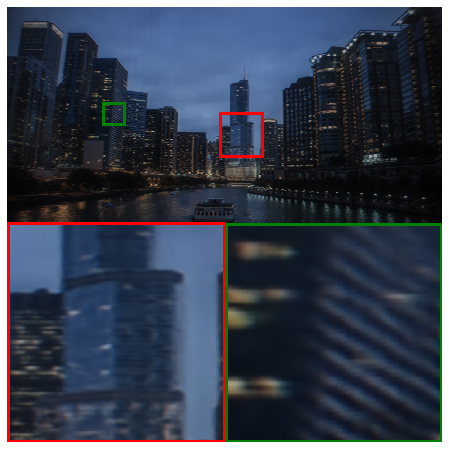}
	\end{subfigure}
	\begin{subfigure}{0.19\linewidth} 
		\includegraphics[width=\textwidth]{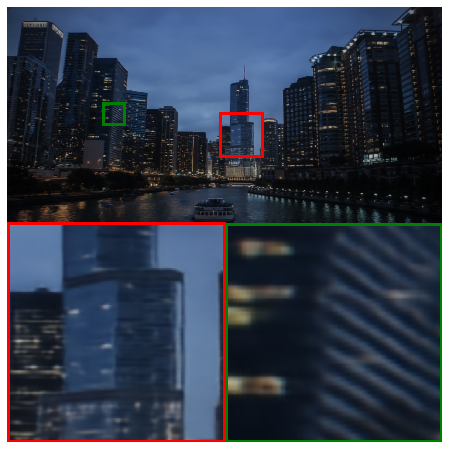}
	\end{subfigure}
	\begin{subfigure}{0.19\linewidth} 
		\includegraphics[width=\textwidth]{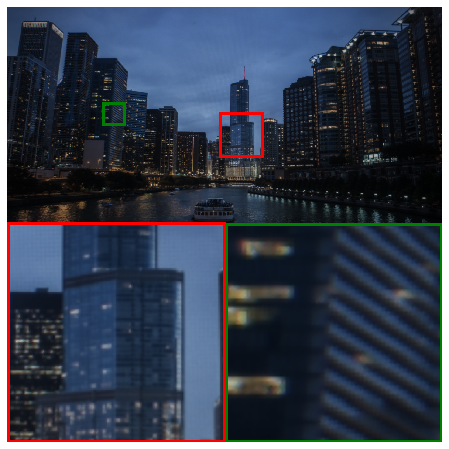}
	\end{subfigure}
	
	\begin{subfigure}{0.19\linewidth} 
		\includegraphics[width=\textwidth]{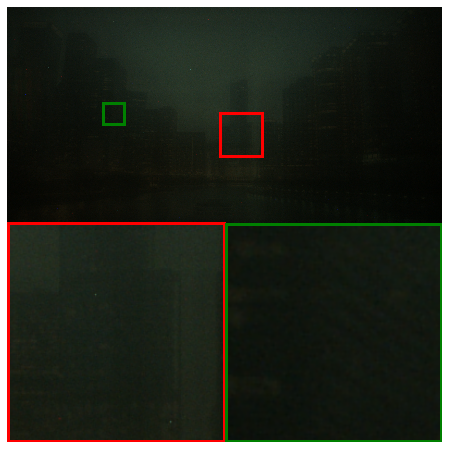}
		\caption{Display} 
	\end{subfigure}
	\begin{subfigure}{0.19\linewidth} 
		\includegraphics[width=\textwidth]{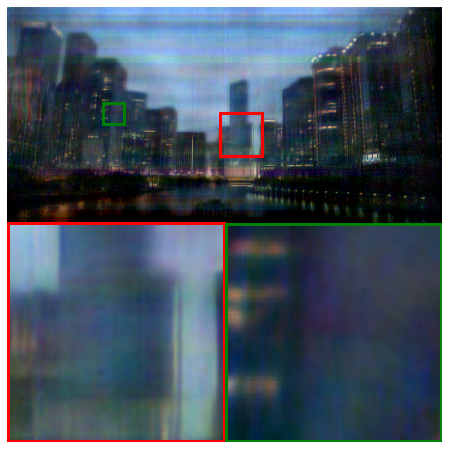}
		\caption{DeP} 
	\end{subfigure}
		\begin{subfigure}{0.19\linewidth} 
		\includegraphics[width=\textwidth]{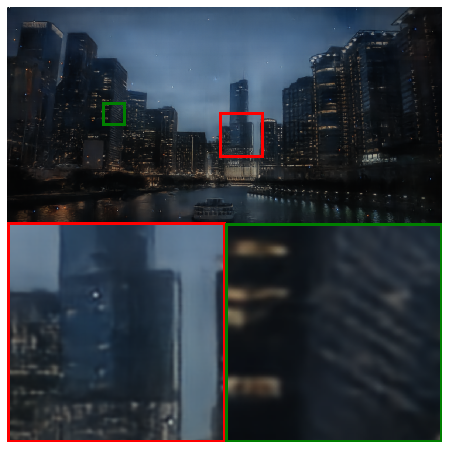}
		\caption{UNet(Syn)} 
	\end{subfigure}
	\begin{subfigure}{0.19\linewidth} 
		\includegraphics[width=\textwidth]{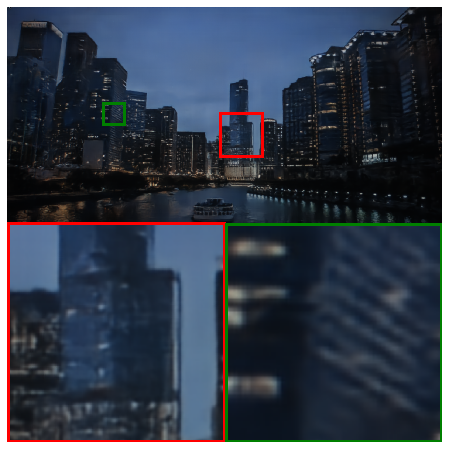}
		\caption{UNet} 
	\end{subfigure}
	\begin{subfigure}{0.19\linewidth} 
		\includegraphics[width=\textwidth]{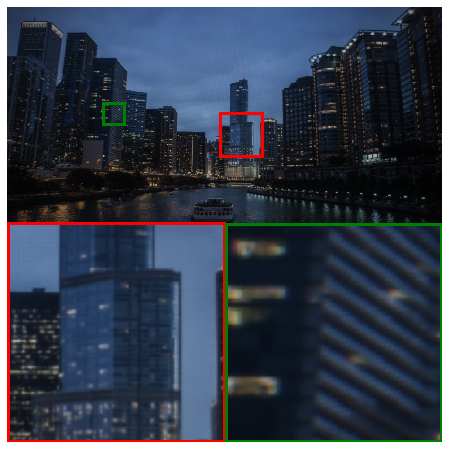}
		\caption{GT} 
	\end{subfigure}
	\caption{More restoration results. For each two-row group, the first row is for T-OLED, and the second one is for P-OLED.}
	\label{fig:result}
\end{figure*}

We show more restoration results in Fig. \ref{fig:result}.

{\small
\bibliographystyle{ieee_fullname}
\bibliography{egbib}
}

\end{document}